\def\arxivversion{1}
\def\withappendix{1}
\ifdefined\pdfminorversion\pdfminorversion=7\fi
\documentclass[10pt,twocolumn,letterpaper]{article}
\ifdefined\arxivversion\usepackage[pagenumbers]{cvpr}\else\usepackage[review]{cvpr}\fi
\usepackage{bm}
\usepackage{multirow}
\usepackage{tabularx}
\usepackage{makecell}
\usepackage{microtype}
\usepackage{placeins}
\usepackage{float}
\usepackage{algorithm}
\usepackage[noend]{algpseudocode}
\usepackage{tikz}
\usetikzlibrary{arrows.meta,positioning,fit,calc,backgrounds,shapes.misc,decorations.pathreplacing}
\definecolor{cvprblue}{rgb}{0.21,0.49,0.74}
\usepackage{xr-hyper}
\usepackage[pagebackref,breaklinks,colorlinks,allcolors=cvprblue]{hyperref}
\def\confName{CVPR}
\def\confYear{2027}
\graphicspath{{figures/}}
\renewcommand{\arraystretch}{1.05}
\newcolumntype{Y}{>{\centering\arraybackslash}X}
\definecolor{obscol}{HTML}{0072B2}   % observation / history
\definecolor{predcol}{HTML}{E69F00}  % prediction / world
\definecolor{actcol}{HTML}{7B61A8}   % action
\definecolor{gtcol}{HTML}{4D4D4D}    % ground truth / neutral
\definecolor{failcol}{HTML}{D55E00}  % failure annotation
\definecolor{fillcolor}{HTML}{C2185B}% placeholders to be filled by the authors

\usepackage{colortbl}
\usepackage{xfp} % \fpeval in \dcell; built into recent LaTeX kernels, needed on older TeX installations
\definecolor{tabbetter}{HTML}{1674AB}
\definecolor{tabworse}{HTML}{C95318}
\newcommand{\bestcell}[1]{\cellcolor{tabbetter!22}\textbf{\boldmath#1}}
\newcommand{\secondcell}[1]{\cellcolor{tabbetter!9}#1}
\newcommand{\dcell}[2]{\edef\dcelltmp{\noexpand\cellcolor{%
  \ifdim #1pt<0pt tabworse\else tabbetter\fi!\fpeval{round(min(abs(#1)/15,1)*40,0)}}}\dcelltmp#2}

\newcommand{\QtwoCase}{1}

\newcommand{\QthreeCase}{3}

\newcommand{\BackboneCase}{1}

\newcommand{\papertitle}{\ours: Role-Aware 4D World--Action Modeling for Dynamic Manipulation}

\newcommand{\oursname}{FoveaWAM}
\newcommand{\ours}{\oursname\xspace}
\newcommand{\oursentry}{\oursname~\tablelang{(Ours)}{（本文）}}
\newcommand{\bench}{\mbox{OnTimeBench}\xspace}
\newcommand{\nativeLead}{20.8}

\newcommand{\cfgVisualEncoder}{MAE ViT-B/16}   % frozen observation encoder (SCALE0: frozen MAE-B selected)
\newcommand{\cfgLayers}{12}
\newcommand{\cfgWidth}{1024}
\newcommand{\cfgHeads}{16}
\newcommand{\cfgTrainableParams}{271.8M}  % 271,825,239; excludes 50 normalization buffers and frozen MAE-B (85,798,656)
\newcommand{\cfgHistoryLen}{8}          % history samples
\newcommand{\cfgHistorySpan}{350\,ms}   % history span
\newcommand{\cfgHorizons}{8}            % number of prediction horizons
\newcommand{\cfgHorizonStep}{50\,ms}

\newcommand{\cfgPatches}{196}           % current-image patch tokens
\newcommand{\cfgMetaTokens}{16}         % learnable summary (meta) tokens of the world stream
\newcommand{\cfgDemos}{800}             % demonstrations per Dynamic-14 task (N800)
\newcommand{\cfgUpdates}{25{,}870}      % actual 10 epochs; 20-epoch cap 51,740; selected update 25,870
\newcommand{\cfgBatch}{512}

\newcommand{\cfgContextSide}{14}
\newcommand{\cfgContextGrid}{$\cfgContextSide\times\cfgContextSide$}

\newcommand{\cfgChunk}{8}               % commands per action chunk
\newcommand{\cfgEuler}{5}               % Euler steps per generated stream
\newcommand{\cfgControlRate}{20\,Hz}
\newcommand{\cfgRobot}{a Flexiv Rizon~4s arm with a GN01 gripper}

\newcommand{\cfgGPU}{RTX~5090}       % robot GPU (32 GB); every method's latency is timed on it at batch size 1 and replayed in simulation
\newcommand{\cfgEvalGPU}{RTX~PRO~6000}  % GPU on which the simulated rollouts are computed; its compute time does not enter the episodes
\newcommand{\resultplaceholder}[1][00.0]{\textcolor{fillcolor}{%
  \begingroup\setbox0=\hbox{#1}\hbox to \wd0{\hfil\textemdash\hfil}\endgroup}}
\newcommand{\Qtwo}[2]{\ifnum\QtwoCase=#1\relax#2\fi}
\newcommand{\Qthree}[2]{\ifnum\QthreeCase=#1\relax#2\fi}
\newcommand{\Bbone}[2]{\ifnum\BackboneCase=#1\relax#2\fi}
\newcommand{\QtwoSel}[3]{\ifcase\QtwoCase\relax\or#1\or#2\or#3\fi}
\newcommand{\QthreeSel}[3]{\ifcase\QthreeCase\relax\or#1\or#2\or#3\fi}
\newcommand{\BboneSel}[2]{\ifcase\BackboneCase\relax\or#1\or#2\fi}
\newcommand{\sel}{$^\star$}
\newcommand{\gridsel}[1]{\ifnum\cfgContextSide=#1\relax\sel\fi}

\newcommand{\mainref}[1]{\ref{#1}}

\newcommand{\R}{\ensuremath{\mathbb{R}}\xspace}
\newcommand{\Exp}{\operatorname{Exp}}

\newcommand{\vect}[1]{\mathbf{#1}}
\newcommand{\tapp}{\tau^{\mathrm a}}

\newcommand{\tablelang}[2]{#1}
\usepackage{amssymb}
\providecommand{\cmark}{\ensuremath{\checkmark}}

\ifdefined\withappendix\else
\fi
\title{\papertitle}
\ifdefined\arxivversion
\author{
Jiaqi Yin$^{1,2}$ \qquad Baoyue Zhang$^{2}$ \qquad Jiyuan Zhang$^{3}$ \qquad Wenxuan Liu$^{2}$\\
Zhaofei Yu$^{2}$ \qquad Yue Deng$^{1,\dagger}$ \qquad Tiejun Huang$^{2,\dagger}$\\[6pt]
$^{1}$Zhongguancun Academy \enspace $^{2}$Peking University \enspace $^{3}$Beijing Academy of Artificial Intelligence\\[2pt]
{\small $^{\dagger}$Corresponding authors}
}
\hypersetup{pdfauthor={Jiaqi Yin, Baoyue Zhang, Jiyuan Zhang, Wenxuan Liu, Zhaofei Yu, Yue Deng, Tiejun Huang}}

  \hypersetup{pdftitle={\papertitle}}
\else
  \author{Anonymous Authors}
  \hypersetup{pdfauthor={Anonymous Authors},pdftitle={FoveaWAM}}
\fi
\begin{document}
\twocolumn[{%
\renewcommand\twocolumn[1][]{#1}%
\maketitle
% Selected real-scene visualization and current Table 1 Native results, 2026-11-04.
% \begin{center}
% \captionsetup{type=figure}
% \input{figures/Fig1/panels}%
% \captionof{figure}{\textbf{\ours allocates prediction precision by role.}
% (a) It forecasts a precise SE(3) trajectory for the target and sparse 3D flow for its
% surroundings. (b) Success rate (SR) at native latency versus p95 time-to-action on \bench:
% \ours has the highest SR at about one tenth the latency of video world--action models.}
% \label{fig:teaser}
% \end{center}

% Official CVPR teaser pattern: the center environment adds the \topsep gap
% between the caption and the two-column body (a bare group adds none).
\begin{center}
\centering
\captionsetup{type=figure}
\resizebox{0.97\linewidth}{!}{%
    % Match the 2260:40:1350 widths of the previous 3650-unit canvas.
\makebox[\textwidth]{%
  \includegraphics[width=.619178\textwidth]{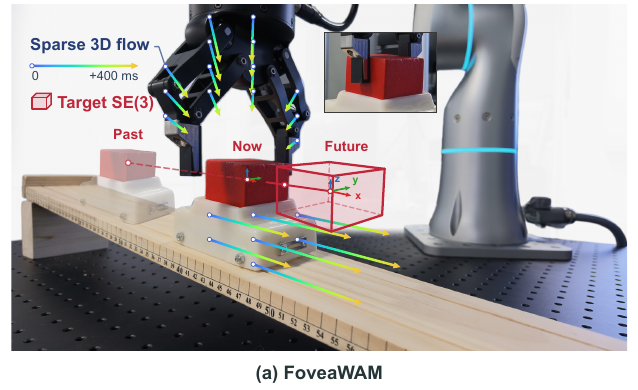}%
  \hfill
  \includegraphics[width=.369863\textwidth]{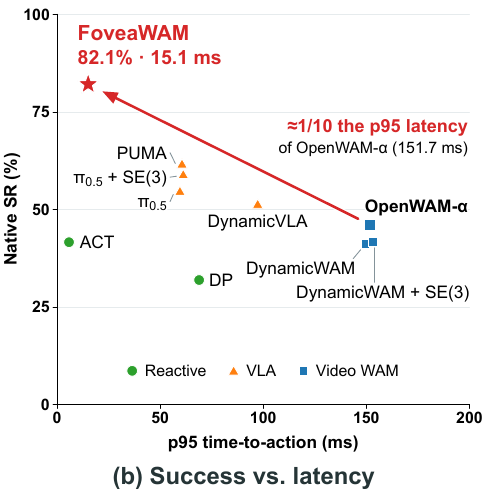}%
}%
}\par
\captionof{figure}{\textbf{\ours allocates prediction precision by role.}
(a) Illustrative target SE(3) poses and sparse 3D flow. The inset shows a staged grasp.
(b) \ours leads in native SR on \bench at about one tenth the p95 latency of OpenWAM-$\alpha$.}
\label{fig:teaser}
\end{center}%
% Fig. 1: full-width teaser under the title (spec 2026-11-02)
}]
\begin{abstract}
To manipulate a moving object, a policy must predict the scene at the moment its action
executes, yet every millisecond spent predicting pushes that moment further away.
Scene-level forecasts are computationally expensive, whereas target-centric forecasts can
omit nearby objects that affect contact. Our key insight is that what a policy
must predict decouples by role: only the target's pose, which determines  contact, requires 
millimeter precision, while the surroundings matter primarily through the clearance they
leave and the motion they impart. We present \ours, a native 4D world--action model
that allocates precision according to these roles. Specifically, it models the target's past and future motion as a
metric SE(3) trajectory, the surroundings as sparse 3D flow, and relies on the current frame
for everything static. This compact representation lets \ours act in 15.1\,ms on one
\cfgGPU, about $10\times$ faster than video world--action baselines, leaving substantially less time for the scene to evolve before action execution.
To evaluate each policy under its own latency, we introduce \bench, a dynamic
manipulation benchmark in which the scene keeps moving while the policy computes.
\ours reaches 82.1\% success on \bench, \nativeLead{} points above the strongest baseline, and
also leads on DOMINO and real-robot tasks.
\par\noindent Project page: \url{https://foveawam.github.io/}.
\end{abstract}
\section{Introduction}
\label{sec:introduction}

% Robot manipulation policies have advanced rapidly in scenes that wait for them
% \cite{diffusionpolicy,pi05,dreamzero}: however long a policy computes, the scene it observed is
% still the scene it acts on. A moving object breaks this assumption. A ball rolling down a ramp
% at about 0.9\,m/s travels 18\,cm during a 200\,ms inference, while contact tolerates only
% millimeters. The policy must therefore act on a prediction of where the scene will be when its action executes. But prediction is itself computation, and it lengthens the very interval it
% must cover: the more a policy predicts, the later it acts, and the further ahead it must
% predict. 

Robot manipulation policies have made substantial progress in static scenes \cite{diffusionpolicy,pi05}.
In these scenes, the world stays still during inference, so the observed scene remains valid when the action executes.
In dynamic manipulation, however, the target keeps moving during inference, so the observed pose is already outdated when the action executes. 
For example, a block sliding down a ramp at about 0.9\,m/s travels 18\,cm during a 200\,ms inference interval, far exceeding the positional tolerance of accurate contact.
Successful manipulation therefore requires predicting the target's pose at execution time. % The policy must also account for surrounding objects, because they may obstruct access or alter the target's motion. Yet prediction itself takes time. A more expensive prediction delays execution and extends the horizon the policy must predict. 
Predicting the target’s motion requires accounting for surrounding objects, but the target continues to move while the prediction is being computed. The longer prediction takes, the further ahead the policy must predict.
This creates a tension between \mbox{accurate prediction and timely action.}

\begin{figure}[t]
\centering
\includegraphics[width=\columnwidth]{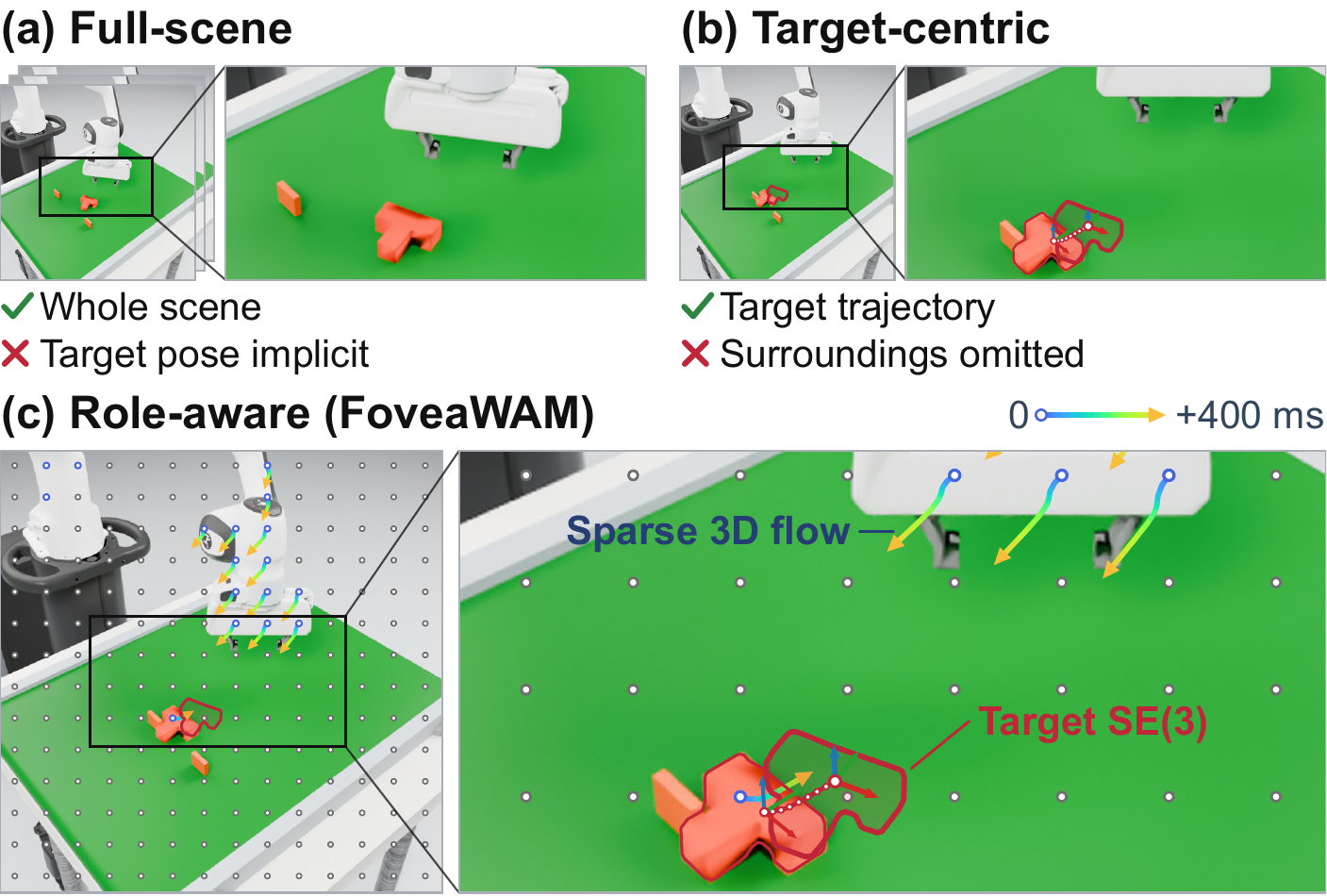}
\caption{\textbf{Where prediction precision goes.}
(a) Full-scene methods, such as video world--action models, predict future
frames of the whole scene and leave the target pose implicit
\cite{dreamzero,openwam_alpha}.
(b) Target-centric methods focus prediction on the object, using future
features \cite{domino} or pose trajectories \cite{spot}.
(c) \ours predicts a precise SE(3) trajectory for the target and only sparse 3D
flow for its surroundings, and reads the static scene from the current frame.}
\label{fig:paradigms}
\end{figure}

Existing methods vary in what they predict (Fig.~\ref{fig:paradigms}).
Scene-level predictors model future images or video \cite{dbctfp,dreamzero,openwam_alpha,dynamicwam}
or 3D motion \cite{physmani}. These predictions capture surrounding motion, but a dense forecast can spend computation on parts of the scene with little bearing
on contact. In a video forecast, the target may occupy only a small region, and its pose remains implicit in the predicted pixels. AHEAD limits its rollout to task-relevant
latent patches \cite{ahead}. PUMA supervises future features of the target \cite{domino}, and SPOT predicts target pose trajectories \cite{spot}. Focusing on the target can omit surrounding objects that block access or alter its motion. For example, a
moving obstacle can close the approach path even if the target follows the
predicted trajectory. Determining whether that path stays clear and estimating
the target's pose at contact impose different demands on a forecast.
\emph{What should a policy predict about a moving scene, and how precisely,
\mbox{to make accurate contact in time?}}

%第二段 existing method 局限Existing methods adopt different prediction strategies, but their prediction scope remains limited (Fig.~\ref{fig}). Full-scene methods predict future videos~\cite{加参考文献}, latents~\cite{加参考文献}, or 3D motion~\cite{加参考文献}, without distinguishing the precision needed for the environment, potentially spending computation on surrounding details unnecessary for contact.\cite{dynamicwam,foretime,openwam_alpha,dreamzero,physmani,dbctfp}分别加 不要挤在一起. Target-centric methods focus on the target's future motion, narrowing the prediction scope but omitting surrounding motion that may obstruct access or alter the target's trajectory \cite{domino,spot}. The former retain more information at greater computational cost, whereas the latter concentrate prediction but lack necessary context. %Thus, the key question raises: \textit{what must be predicted, and at what precision, to make accurate contact in time?}

% Our key observation is that prediction requirements \textbf{differ by role} (Fig.~\ref{fig:teaser}). 
% The target's position and orientation at execution time directly determine contact success and therefore require precise prediction. 
% In contrast, the surroundings primarily constrain access and influence target motion, so their motion can often be represented coarsely, while static context is available in the current observation.
% A policy can therefore concentrate precise prediction on the target while retaining necessary environmental information at lower cost. This resembles the division of labor in biological vision: the fovea provides fine detail, while peripheral vision preserves broad motion awareness~\cite{curcio1990photoreceptor,strasburger2011peripheral,rosenholtz2016peripheral}.

Our key insight is that prediction requirements \textbf{differ by role} (Fig.~\ref{fig:teaser}). 
The target's position and orientation at execution time directly determine contact success and therefore require precise prediction. 
In contrast, the surroundings affect the contact only indirectly, by obstructing
access or altering the target's motion, so their motion requires only coarse
prediction. 
Finally, the rest of the scene is static, so it is already available in the current frame and requires no prediction. 
A policy can therefore make accurate contact in time by concentrating precise prediction on
the target and capturing the surroundings at low cost. 
This allocation resembles biological vision, where the fovea resolves fine detail and
peripheral vision preserves
\mbox{broad motion awareness~\cite{curcio1990photoreceptor,strasburger2011peripheral,rosenholtz2016peripheral}.}

Based on this insight, we propose \textbf{\ours}, a native 4D world--action model that allocates precision by role (Fig.~\ref{fig:paradigms}c). 
\ours observes and predicts scene motion directly in metric 3D space over time, rather than in pixels. 
In this space, each part of the scene can be represented separately, so each receives the precision that its role requires. 
The target is predicted precisely, as a metric SE(3) trajectory that gives its future position and orientation explicitly with only a few parameters. 
In contrast, the surroundings are predicted coarsely, as sparse 3D flow, and the static scene is not predicted. 
The same allocation applies to the input. 
A single frame already shows the static scene and the layout of the surroundings, so only the target's motion requires a history. 
\ours therefore reads one RGB-D frame and a short SE(3) pose history of the target, instead of the image or optical-flow
histories used by existing dynamic policies~\cite{dynamicvla,domino,dynamicwam}.
Since both the input and the predictions are compact, \ours acts in 15.1\,ms on one \cfgGPU{}, about $10\times$ faster than video world--action models, and the scene has little time to move before the action executes.

The tension between accurate prediction and timely action can be measured only when the evaluation accounts for inference time. However, the default evaluators of DOMINO \cite{domino} and PhysMani-Bench \cite{physmani} pause the simulation during inference, so a slow policy is scored as if it were instant. 
We therefore introduce \textbf{\bench}, a benchmark of 14 dynamic manipulation tasks in which the world never pauses and every policy is evaluated at its native latency. 
On the same scenes, pausing the world overestimates slow policies and changes the ranking of methods (Fig.~\ref{fig:latency}c).

Experiments show that \ours makes accurate contact in time. 
On \bench, it reaches 82.1\% success at native latency, \nativeLead{} points above the strongest baseline.
It also leads by 19.9 points on the single-arm tasks of DOMINO \cite{domino}, whose evaluator pauses during inference, so its advantage does not come from speed alone. 
On six real-robot tasks, it achieves the highest success on every task and 48.3\% on average, 17.5 points above the strongest baseline. 
Finally, ablations show that each part of the role-aware design contributes to success,
with the largest drop from removing the target's pose history.

% On \bench, \ours reaches 82.1\% success at native latency, \nativeLead{} points above the strongest baseline.
% It also leads by 19.9 points on DOMINO. 
% Since the DOMINO evaluator pauses during inference, this gain comes from the representation and not from speed. 
% On six real-robot tasks, \ours achieves the highest success on every task and 48.3\% on average, 17.5 points above the strongest baseline.
% Ablations confirm that each part of the allocation contributes. 
% Replacing the target's SE(3) trajectory with the same sparse flow as its surroundings, removing the sparse flow, or removing the pose history lowers success by 4.5, 4.2, and 19.6 points, respectively.

Our key contributions are as follows:
\begin{itemize}[leftmargin=*,topsep=2pt,itemsep=1pt]
\item We identify distinct prediction requirements in dynamic manipulation and
propose a role-aware prediction paradigm. It forecasts target motion precisely
and surrounding motion coarsely, using the current frame for static geometry.
This allocation preserves the precision needed for contact while reducing
prediction cost.
\item We present \textbf{\ours}, a native 4D world--action model that
instantiates this paradigm and achieves precise, low-latency action generation.
It acts in 15.1\,ms, about $10\times$ faster than video world--action models,
and outperforms the strongest baseline by \nativeLead{} points on \bench. It also leads
on DOMINO's single-arm tasks with paused physics and across six dynamic real-robot tasks.
\item We build \textbf{\bench}, a benchmark of 14 dynamic manipulation tasks in
which the world never pauses and every policy is evaluated at its native
latency, exposing the cost of latency that paused evaluation hides. 
\end{itemize}
%贡献1 一个建立在role aware发现得新范式。这个范式突破什么什么瓶颈，为推动xxx建立一个新基准。We identify distinct prediction requirements for the target and its surroundings in dynamic contact and introduce a role-aware prediction paradigm. By allocating representations and precision according to role, it alleviates the tension between accurate prediction, environmental context, and timely execution.
%贡献2 双流模型设计。 将连续目标位姿历史与当前观测相结合，利用世界—动作双流架构复用世界上下文，实现精确、低延迟的动作生成。
%贡献3 bench

\section{Related Work}
\label{sec:related_work}

\paragraph{Dynamic manipulation.}
Earlier systems used explicit tracking and interception planning for moving objects
\cite{allen1993,baeuml2010,marturi2019dynamic}. Recent methods learn to handle motion
from data \cite{dynawm,dynamicmanip,dynamicvla}. They use optical-flow histories
\cite{domino,dynamicwam} or condition actions on forecasts at inference
\cite{foretime,ahead,physmani}. \ours encodes past motion with a metric SE(3) pose
history and a single current frame. Asynchronous execution and chunk-transition
methods address action timing and continuity \cite{bid,rtc,smolvla,vlash,dynamicvla}.
The default DOMINO \cite{domino} and PhysMani-Bench \cite{physmani} rollouts step
physics after actions return. \bench advances physics during inference to evaluate
each policy at its native latency.

\begin{figure*}[t]
\centering
\includegraphics[width=\textwidth]{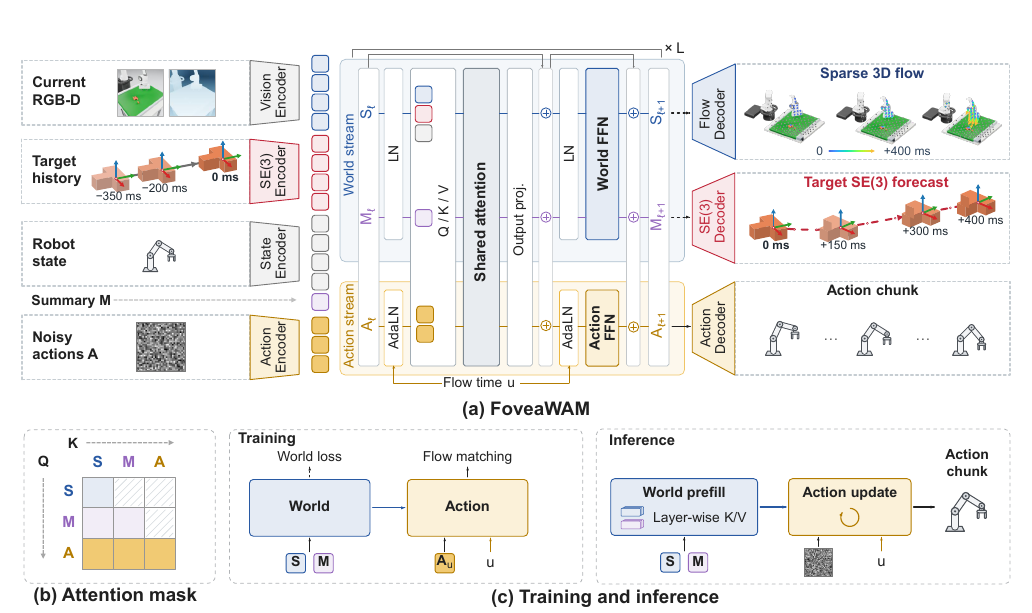}
\caption{\textbf{Overview of \ours.}
(a) Architecture. The current RGB-D frame and the target and robot histories form scene tokens $S$, alongside summary tokens $M$ and noisy action tokens $A$. All tokens share masked attention, followed by separate world and action FFNs. During training, linear decoders predict the sparse 3D flow from $S$ and the target SE(3) trajectory from $M$.
(b) Attention mask. Rows are queries, columns are keys, and hatched cells are masked.
(c) Training and inference. The two streams are trained jointly. At inference, the decoders are skipped, and the world stream runs once per query and caches its keys and values for the action stream.}
\label{fig:architecture}
\end{figure*}

\paragraph{Forecasting in world--action models.}
World--action models (WAMs) couple video and action prediction \cite{uwm,uva,dreamzero,openwam_alpha},
building on video prediction as a policy objective \cite{gr1,seer}. Recent models
predict features or latent tokens \cite{flare,dreamvla,wog}, masks \cite{maskwam}, flow
\cite{dreamwam}, or object slots \cite{oawam} alongside or in place of RGB. PUMA,
Fast-WAM, and SG-WAM use future supervision during training and can omit the forecast
at deployment \cite{domino,fastwam,sgwam}. During training, \ours supervises motion in
a role-aware metric 4D state that explicitly represents the target's future pose.

\paragraph{Object-centric and geometric motion representations.}
Manipulation policies use metric 3D inputs \cite{peract,dp3}. Related representations
include point tracks \cite{atm,track2act,im2flow2act}, 3D flow
\cite{generalflow,flowdreamer,pointworld,physmani}, object keypoints or trajectories
\cite{kpam,spot,objectforesight}, and rigid-body decompositions \cite{se3nets,se3posenets}.
DOCIR separates robot, target, and obstacles in visual inputs \cite{docir}. Gaze-based
robot perception concentrates sensing through foveated inputs
\cite{kim2021gaze,chuang2026lookfocusact} or active camera control \cite{kerr2025eyerobot}.
\ours allocates prediction precision by role, using coarse sparse flow for the scene
and \mbox{a metric SE(3) trajectory for the target.}

\section{\ours}
\label{sec:method}

\ours applies the role-aware allocation to both what it reads and what it predicts
(Fig.~\ref{fig:architecture}a). It reads the current RGB-D frame for the static scene
and the context, and an SE(3) pose history for the target. It predicts coarse sparse 3D
flow for the scene and a precise SE(3) trajectory for the target. In between, a world
stream (scene tokens $S$ and summary tokens $M$) and an action stream (noisy action
tokens $A$) share one masked attention in each of the $L$ layers.

\subsection{Problem Formulation}
\label{sec:formulation}

We study manipulation in scenes that keep moving while the policy computes. At query
$t$, the policy $\pi$ reads an RGB-D frame $(I_t,D_t)$ captured at time $\tau_t$, a
history $\mathcal T_t$ of the target pose $T(\tau)=(R(\tau),\vect p(\tau))\in\mathrm{SE}(3)$,
and a history $\mathcal E_t$ of the robot state, all in the robot base frame. Learned
from demonstrations, the policy outputs a chunk of $H$ commands at control period
$\Delta_a$,
\begin{equation}
(\vect a_{t,0},\ldots,\vect a_{t,H-1})\sim\pi\big(\cdot\mid I_t,D_t,\mathcal T_t,\mathcal E_t\big),
\label{eq:policy}
\end{equation}
where command $j$ executes at $\tau_t+j\Delta_a$.

Each task assigns the scene three roles. The \emph{target} is the object whose pose the
robot must match within a contact or containment tolerance. The \emph{context} is the
part of the surroundings that affects contact. It contains the obstacles that the robot
must clear and the drivers that move the target. Everything else is \emph{static}. The
chunk becomes available only after inference, so each executed command must account for
the target and the context at its execution time.

\subsection{Role-Aware Dynamics Encoding}
\label{sec:inputs}

A single frame shows where things are but not how they move. The context does not need
a motion history of its own. The clearance it leaves is visible in the current geometry,
and the motion it imparts appears in how the target has moved. \ours therefore pairs one
RGB-D frame with a short metric pose history of the target.

\paragraph{Scene tokens from the current frame.}
Contact is decided in metric space, so \ours anchors each patch token at a metric 3D
point. A frozen \cfgVisualEncoder{} \cite{mae} encoder turns the RGB image into
\cfgPatches{} patch tokens on a \cfgContextGrid{} grid. The depth map back-projects the
center of patch $i$ to a point $\vect x_i\in\R^3$ in the base frame. A linear embedding
of $\vect x_i$ is added to the patch token, and $\vect x_i$ also sets the 3D rotary
position embeddings of its queries and keys. Attention between patches then depends on
their relative metric positions, and observations, predictions, and actions share one
frame.

\paragraph{Target tokens from the pose history.}
The pose history covers the last \cfgHistorySpan{} in \cfgHistoryLen{} samples at
\cfgControlRate{}. The SE(3) encoder turns it into \cfgHistoryLen{} tokens, one for the
current target pose and seven for the motions between consecutive samples. Each motion
token holds the translation, relative rotation, time interval, and finite-difference
velocity and acceleration in physical units, so the quantities that decide where the
target will be enter the model directly. The state encoder adds \cfgHistoryLen{} robot
tokens with the end-effector states at the same times. The whole history thus
costs 16 low-dimensional tokens, whereas each additional frame would cost
\cfgPatches{} patch tokens. The patch, target, and robot tokens together form the scene
tokens $S$. The target poses come from the simulator in simulation and from an online
RGB-D tracker on the robot.

\subsection{Role-Aware 4D Future Prediction}
\label{sec:predict}
\begin{figure*}[t]
\centering
\includegraphics[width=\textwidth]{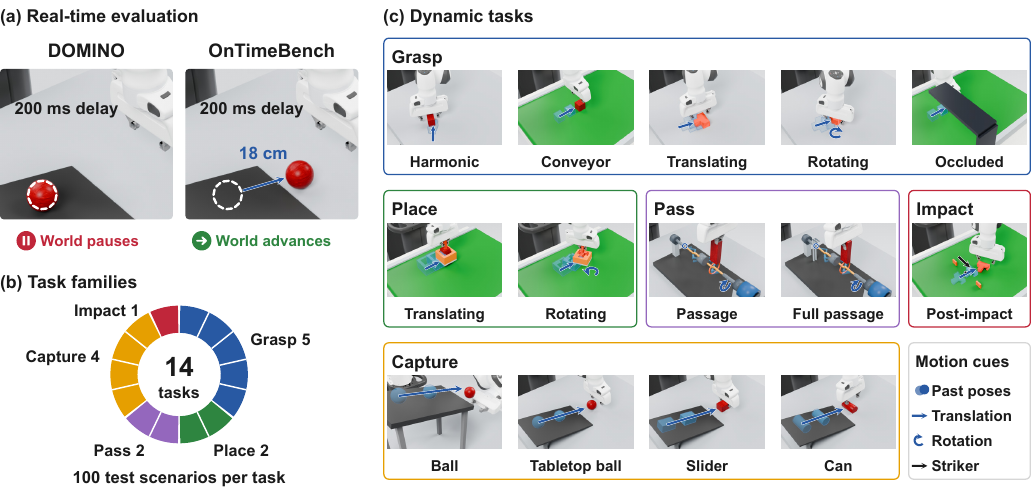}
\caption{\textbf{Overview of \bench.} \bench evaluates each policy at its native
latency on 14 dynamic tasks in five families, with 1{,}400 held-out test scenes.
(a) During a 200\,ms inference, the default DOMINO evaluator~\cite{domino} pauses the
world, while \bench keeps it moving. (b) Tasks per family. (c) The
14 tasks, with motion cues overlaid.}
\label{fig:bench_overview}
\end{figure*}
% figure*: placed here so it lands on the page where Sec. 4 starts

Each command acts at a future time, so the world stream must also represent the scene
at the execution times. The target's pose decides contact, so the SE(3) decoder predicts
its full trajectory. The context matters through clearance and imparted motion, which a
coarse flow already captures. Both predictions are base-frame changes from the current
state at $\tau_t+\delta_h$, with $\delta_h=h\cdot$\cfgHorizonStep{} for
$h=1,\ldots,\cfgHorizons$, and together with the current state they cover the execution
times of the whole chunk. Both decoders are linear and run only during training. A
linear readout forces the world-stream states themselves to encode the predicted motion.

\paragraph{Sparse 3D scene flow.}
The flow covers every surface that the patches see, including the robot, the target,
and the context. For each patch point $\vect x_i$ and horizon $h$, the flow decoder reads
the final state of patch token $i$ and predicts the displacement $\hat F_{ih}\in\R^3$ and
a visibility logit $\hat o_{ih}$, trained with
\begin{equation}
\mathcal L_F=\frac1{|\mathcal V_F|}\sum_{(i,h)\in\mathcal V_F}\!
\big[\|\hat F_{ih}-F_{ih}\|^2+\mathrm{BCE}(\hat o_{ih},o_{ih})\big],
\label{eq:flowloss}
\end{equation}
where $F_{ih}$ and $o_{ih}$ are the true displacement and visibility at $\vect x_i$, and
$\mathcal V_F$ contains the valid pairs. Displacements are divided by a scale $s$ fitted
on the training labels, and the squared error is averaged over coordinates.

\paragraph{Target SE(3) trajectory.}
The flow samples the target only at the few patches it covers, which is too sparse to
determine its rotation. The world stream therefore also holds \cfgMetaTokens{} learned
summary tokens $M$ that attend to the whole scene, and the SE(3) decoder predicts the
target's rigid motion from their mean final state. For each horizon $h$, it predicts a displacement
$\hat{\vect d}_h$ and a rotation increment $\hat{\boldsymbol\phi}_h$, trained with
\begin{equation}
\mathcal L_P=\frac1{|\mathcal V|}\sum_{h\in\mathcal V}\Big[\|\hat{\vect d}_h-\vect d_h\|^2
+\lambda_R\,d_{\mathrm{SO}(3)}\big(\hat R_h,R_h\big)^2\Big],
\label{eq:poseloss}
\end{equation}
where $\vect d_h=\vect p(\tau_t{+}\delta_h)-\vect p(\tau_t)$ and
$R_h=R(\tau_t{+}\delta_h)$ are the true displacement and orientation,
$\hat R_h=\Exp(\hat{\boldsymbol\phi}_h^\wedge)R(\tau_t)$, $d_{\mathrm{SO}(3)}$ is the
geodesic distance, and $\mathcal V$ contains the horizons with valid labels.
Rotation matters for contact. An error $\theta$ moves a contact point at distance $r$
from the target's origin by up to $2r\sin(\theta/2)$, nearly 9\,mm for $\theta=5^\circ$
and $r=10$\,cm.

\subsection{World--Action Learning and Inference}
\label{sec:act}

\paragraph{Action stream.}
The action stream holds $H=\cfgChunk$ tokens. Each token embeds one noisy command, which
consists of an end-effector translation increment, a rotation increment, and a gripper
command. World tokens use layer normalization, and
action tokens use adaptive layer normalization conditioned on the flow time $u$. The
attention mask lets $S$ attend to $S$, $M$ to $S$ and $M$, and $A$ to all tokens
(Fig.~\ref{fig:architecture}b). In every layer, the action stream reads the patch
states shaped by the scene flow and the summary states shaped by the target trajectory.
The world stream never reads the actions, so its states do not depend on the noise or
flow time.

\paragraph{Training objective.}
We train the action stream $v_\psi$ with flow matching \cite{flowmatching}.
With the normalized chunk $\vect a_1$, noise $\vect a_0\sim\mathcal N(0,I)$, and
$u\sim\mathcal U[0,1]$, it regresses the velocity
\begin{equation}
\mathcal L_A=\mathbb E\,\mathrm{MSE}\big(v_\psi(\vect a_u,u\mid\mathcal S_t),\,\vect a_1-\vect a_0\big),
\label{eq:fm}
\end{equation}
where $\vect a_u=(1-u)\vect a_0+u\vect a_1$, $\mathcal S_t$ denotes the inputs in
Eq.~\eqref{eq:policy}, and the MSE averages over valid commands and action dimensions. Both streams are trained jointly with
\begin{equation}
\mathcal L=\mathcal L_A+\lambda_F\mathcal L_F+\lambda_P\mathcal L_P+\lambda_{\rm aux}\mathcal L_{\rm aux},
\label{eq:total}
\end{equation}
where $\mathcal L_{\rm aux}$ sums auxiliary appearance losses.

\paragraph{Inference.}
\label{sec:inference}
At deployment, \ours skips the flow and SE(3) decoders, as in the action-only inference
of prior video WAMs \cite{uva,fastwam} (Fig.~\ref{fig:architecture}c). Since the world states
do not depend on the noise, the world stream runs once per query and caches its keys
and values. The action stream then integrates \cfgEuler{} Euler steps
on this cache. One query takes 15.1\,ms at p95 on one \cfgGPU{}, within one control
period. Commands are indexed by time from capture, so the policy needs no latency input
(Sec.~\ref{sec:bench_protocol}).

\section{\bench}
\label{sec:bench}

The default evaluators of DOMINO \cite{domino} and PhysMani-Bench \cite{physmani}
advance the simulation only after the policy returns its actions, so inference takes
no time in the scene.
A policy that computes for 200\,ms thus still finds the ball where it observed it,
although a real ball would have rolled 18\,cm (Fig.~\ref{fig:bench_overview}a).
In \bench, observation, scene motion, and action execution share one physical clock.
The scene keeps moving while the policy computes, so each policy is evaluated at its
native latency.

\bench contains 14 tasks in five families (Fig.~\ref{fig:bench_overview}b, c).
Grasp tasks pick up a moving target, and Place tasks put a held object into a moving
receptacle.
Pass tasks guide a wand along a rotating wire without contact, Capture tasks catch a
rolling or sliding object, and the Impact task grasps a part after a striker hits it.
Target speeds range from 0.03 to 1.14\,m/s.
The tasks also differ in what decides success.
In the seven rotation-sensitive tasks, the target rotates and its orientation matters
at contact.
In the nine context-active tasks, an obstacle limits the clearance or a driver such as
a conveyor, ramp, or striker moves the target.

\subsection{Construction}
\label{sec:bench_construction}
\bench is built in MuJoCo and robosuite \cite{mujoco,robosuite} with a Franka Panda arm
controlled at \cfgControlRate{}.
Each scene samples the target's initial state and motion parameters within
task-specific ranges and is kept only if it is physically feasible.
A task-specific expert with privileged simulator state provides \cfgDemos{}
demonstrations per task.
A demonstration is kept only if the expert succeeds, a replay of its actions reproduces
the episode, and its path and jerk pass quality checks.
We select configurations on 1{,}400 development scenes and report all \bench results on
1{,}400 held-out test scenes, 100 per task, shared by all methods.

\subsection{Evaluation Protocol}
\label{sec:bench_protocol}
\begin{figure}[t]
\centering
\includegraphics[width=\columnwidth]{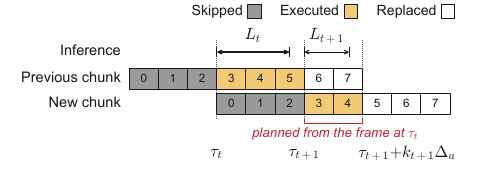}
\caption{\textbf{Native-latency execution.} Query $t$ captures a frame at $\tau_t$
and runs for $L_t$ while the world keeps moving. Its chunk takes over at the next
control boundary $\tau_{t+1}=\tau_t+k_t\Delta_a$, with $k_t=\lceil L_t/\Delta_a\rceil$,
where the next query captures. Overdue commands before $k_t$ are skipped, and the
previous chunk runs until then.}
\label{fig:protocol}
\end{figure}
In \emph{Native}, physics advances during each query by the method's latency,
drawn per query from its distribution measured at batch size 1 on one \cfgGPU{}.
All methods share one executor, which hands over each chunk at the next control
boundary (Fig.~\ref{fig:protocol}).
\emph{+150\,ms} adds 150\,ms to the native latency of every method, and \emph{Paused}
pauses the world during inference.
We report the task-balanced success rate (SR) and the p95 time-to-action, the time from
capturing an observation until its first command is available.
Task definitions and the full protocol are in the supplementary material.

\section{Experiments}
\label{sec:experiments}

We evaluate \ours on \bench, on DOMINO \cite{domino}, and on a real robot. On \bench,
the paused, native, and delayed settings separate how well a policy represents the
target's motion from how quickly it acts.

\subsection{Implementation Details}
\label{sec:setup}

\paragraph{Baselines.}
We compare with the reactive policies ACT \cite{act} and Diffusion Policy (DP)
\cite{diffusionpolicy}, the vision--language--action model (VLA) $\pi_{0.5}$ \cite{pi05}, the dynamic VLAs DynamicVLA
\cite{dynamicvla} and PUMA \cite{domino}, the WAM OpenWAM-$\alpha$ \cite{openwam_alpha},
and the dynamic WAM DynamicWAM \cite{dynamicwam}. Each baseline keeps its official
architecture and training recipe and is trained on the same demonstrations. To isolate
the effect of target-motion input, $\pi_{0.5}$ and DynamicWAM also run with our SE(3)
history tokens (+SE(3)). OpenWAM-$\alpha$ also runs without its official caching and
compilation (off).

\paragraph{Model and training.}
The \ours transformer has \cfgLayers{} layers of width \cfgWidth{} with \cfgHeads{}
heads and \cfgTrainableParams{} trainable parameters on top of the frozen
\cfgVisualEncoder{}. On \bench, it is trained from scratch on the agent view for 10 epochs (\cfgUpdates{} updates) and runs in compiled BF16.

% Transcribed from results/realtime_c/table1-20261104.json: eleven configurations; all three BF16 Full last protocols are accepted.
% Highlight reported values per column with \bestcell/\secondcell (SR: highest; p95: lowest); exclude pending cells.
\begin{table}[t]
\centering\footnotesize
\caption{\tablelang{\textbf{Quantitative comparison on \bench.} SR (\%) at native latency, with 150\,ms added, and with the world paused. p95: time-to-action on one \cfgGPU{}. Baselines use only the main view. Best and second best are highlighted.}{\textbf{\bench 上的定量比较。}各方法在原生延迟、附加 150\,ms 以及推理期间暂停世界三种设置下的 SR（\%）。p95：在一块 \cfgGPU{} 上的出动作时间。表中基线成功率来自单主视角评测。历史：过去的 RGB 帧、光流或目标 SE(3) 位姿。off：关闭缓存与编译。每列的最佳与次佳已标出。}}
\label{tab:main}
\setlength{\tabcolsep}{1.3pt}
% Equal-width result columns retain right padding for the cell colors.
\begin{tabularx}{\linewidth}{@{}lccc*{4}{Y}}
\toprule
 & \multicolumn{3}{c}{\tablelang{History}{历史}} & \multicolumn{3}{c}{SR (\%) $\uparrow$} & \\
\cmidrule(lr){2-4}\cmidrule(lr){5-7}
\tablelang{Method}{方法} & RGB & \tablelang{Flow}{光流} & SE(3) & \tablelang{Native}{原生} & \makecell{+150\\ms} & \tablelang{Paused}{暂停} & \makecell{p95\\(ms)$\downarrow$} \\
\midrule
ACT \cite{act} &  &  &  & 41.6 & 24.6 & 40.6 & \bestcell{5.8} \\
DP \cite{diffusionpolicy} & \cmark &  &  & 31.9 & 16.1 & 32.0 & 68.9 \\
$\pi_{0.5}$ \cite{pi05} &  &  &  & 54.4 & 28.1 & 66.6 & 59.6 \\
$\pi_{0.5}$+SE(3) &  &  & \cmark & 58.7 & 28.0 & \secondcell{76.0} & 61.2 \\
\midrule
DynamicVLA \cite{dynamicvla} & \cmark &  &  & 51.1 & 23.5 & 54.9 & 97.3 \\
PUMA \cite{domino} &  & \cmark &  & \secondcell{61.4} & \secondcell{31.1} & 53.1 & 60.6 \\
\midrule
OpenWAM-$\alpha$ \cite{openwam_alpha} &  &  &  & 46.1 & 24.8 & 58.7 & 151.7 \\
OpenWAM-$\alpha$ (off) &  &  &  & 15.8 & 10.4 & 58.4 & 466.3 \\
\midrule
DynamicWAM \cite{dynamicwam} &  & \cmark &  & 41.1 & 20.6 & 49.6 & 149.6 \\
DynamicWAM+SE(3) &  & \cmark & \cmark & 41.7 & 24.9 & 64.0 & 153.3 \\
\midrule
\oursentry &  &  & \cmark & \bestcell{82.1} & \bestcell{34.1} & \bestcell{90.3} & \secondcell{15.1} \\
\bottomrule
\end{tabularx}
\end{table}

\begin{figure}[t]
\centering
\definecolor{rankup}{HTML}{1B9E77}\definecolor{rankdown}{HTML}{B5294E}% Rank colors in (c)
\includegraphics[width=\columnwidth]{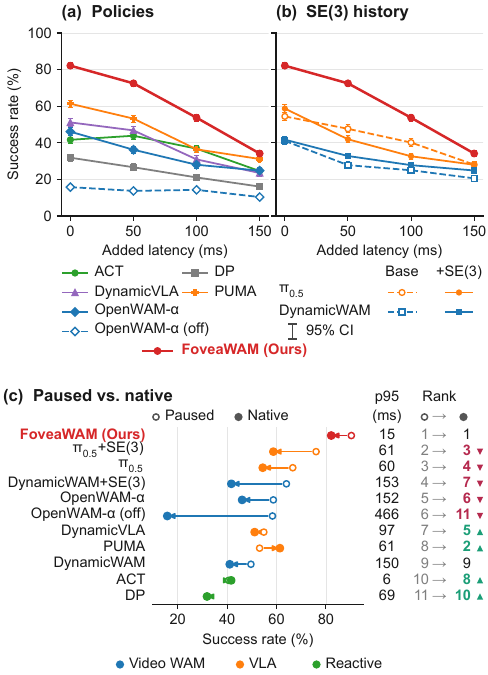}
\caption{\textbf{Latency on \bench.} (a, b) SR versus added delay, for all policies (a) and for base versus +SE(3) variants (b), with 95\% CIs. (c) Paused
and native SR, colored by class as in Fig.~\ref{fig:teaser}, with rank changes
({\color{rankup}$\blacktriangle$}/{\color{rankdown}$\blacktriangledown$}).}
\label{fig:latency}
\end{figure}

\subsection{Dynamic Manipulation Results}
\label{sec:main_results}

\paragraph{Results on \bench.}
As shown in Table~\ref{tab:main}, \ours achieves the highest SR in all three settings.
At native latency, it reaches 82.1\%, \nativeLead{} points above the strongest baseline, PUMA
(61.4\%), and it has the highest or tied-highest SR on 13 of the 14 tasks, as
reported in the supplementary material. The margins are largest where a moving receptacle
or the target's orientation decides success. They reach 30 and 55 points on placement
into translating and rotating receptacles and 28 points on full wire passage. This
lead comes from the role-aware design. Given the same SE(3) history as input,
$\pi_{0.5}$ and DynamicWAM reach 58.7\% and 41.7\%, which is 23.4 and 40.4 points below
\ours.

\paragraph{Representation and latency.}
\ours leads both in how it represents the target's motion and in how quickly it acts.
With the world paused, inference costs no time, and \ours
still leads $\pi_{0.5}$+SE(3), which reads the same target poses, by 14.3 points. This
gap reflects the representation alone. At native latency, \ours loses 8.1 points and
$\pi_{0.5}$+SE(3) loses 17.3, so the gap widens to 23.4 points. Low latency alone is not
enough. ACT is the only method faster than \ours, but it reaches 41.6\%, which is 40.5
points below \ours. \ours leads at every added delay (Fig.~\ref{fig:latency}a). With 150\,ms
added, each chunk must cover a longer open-loop interval. Every method falls to
35\% or below, and the margin over PUMA narrows to 3.0 points.

% Table 2 is input here so that it floats next to the DOMINO results.
% Transcribed from results/domino_baselines/w5-formal-20261111.json (baselines) and results/domino_baselines/foveawam-epoch24-scenes.csv (FoveaWAM, step 2520).
\begin{table}[t]
\centering\footnotesize
\caption{\tablelang{\textbf{Quantitative comparison on DOMINO} \cite{domino}. SR (\%) on its 24 single-arm tasks, 8 Intercept and 16 Track. MS: manipulation score. Best and second best are highlighted.}{\textbf{DOMINO \cite{domino} 上的定量比较。}其 24 个单臂任务（8 个拦截、16 个跟踪）上的 SR（\%）。MS：操作得分。最佳与次佳已标出。}}
\label{tab:external}
\setlength{\tabcolsep}{2.2pt}
% Keep right padding so the MS highlight ends at the rule.
\begin{tabularx}{\linewidth}{@{}l*{4}{Y}}
\toprule
\tablelang{Method}{方法} & \tablelang{Intercept}{拦截} & \tablelang{Track}{跟踪} & \tablelang{All}{全部} $\uparrow$ & MS $\uparrow$ \\
\midrule
$\pi_{0.5}$ & 23.8 & 8.1 & 13.3 & 27.8 \\
DynamicWAM & 16.8 & \secondcell{31.9} & \secondcell{26.9} & \secondcell{41.4} \\
PUMA & \bestcell{36.3} & 12.2 & 20.2 & 35.4 \\
\oursentry & \secondcell{29.4} & \bestcell{55.5} & \bestcell{46.8} & \bestcell{62.0} \\
\bottomrule
\end{tabularx}
\end{table}

\paragraph{Motion input under latency.}
Precise motion input helps only when the action arrives in time
(Fig.~\ref{fig:latency}b). The SE(3) history raises $\pi_{0.5}$ by 9.4 points with the
world paused but by only 4.3 points at native latency. With 50 and 100\,ms added, it
lowers the SR of $\pi_{0.5}$ by 5.7 and 7.5 points, and the 95\% intervals do not
overlap. The history describes the target at capture time, so its value decays as the
action is delayed. \ours is designed to keep this value. It encodes the history
in 16 tokens and acts within one control period, so the history is still current when
the action executes. The history therefore adds 19.6 points to \ours at native latency,
against 4.3 points for $\pi_{0.5}$ (Table~\ref{tab:analysis}).

\paragraph{Paused evaluation.}
Pausing the world hides the cost of latency and reorders 9 of the 11 methods
(Fig.~\ref{fig:latency}c). The two OpenWAM-$\alpha$ deployments share one policy and
reach 58.7\% and 58.4\% with the world paused. At native latency, with p95 of 151.7 and
466.3\,ms, they reach 46.1\% and 15.8\%. The three slowest methods all lose
rank, and the strongest baseline changes from $\pi_{0.5}$+SE(3) to PUMA.

\paragraph{Results on DOMINO.}
DOMINO \cite{domino} has its own tasks, demonstrations, and evaluator. We train \ours,
$\pi_{0.5}$, DynamicWAM, and PUMA on its 24 single-arm Franka tasks. \ours achieves
46.8\% SR and 62.0 MS, 19.9 and 20.6 points above the strongest baseline, DynamicWAM
(Table~\ref{tab:external}). The evaluator pauses physics during inference, so latency
brings no advantage. This lead therefore shows the value of the role-aware
representation itself. With its lead in the paused setting of \bench, \ours leads
whether or not latency is counted. The margin is largest on the 16
Track tasks, where the robot must keep a consistent pose relative to the moving target.
The SE(3) trajectory describes the target's future poses directly, and \ours reaches
55.5\% against 31.9\%. PUMA leads on the 8 Intercept tasks, which require a
single contact with the target (36.3\% against 29.4\%).

\subsection{Ablation Study}
\label{sec:principles}
Table~\ref{tab:analysis} ablates each part of the role-aware design. All
variants except \emph{w/o history} change only the training supervision and run at the
latency of \ours.

% Table 3 is input after the section start so that it floats to the top of the next column.
% Transcribed from se3-wam-focused/research/evidence/hard/ablation-bf16/checkpoint-matrix64.json (latest.pt for each model).
% Rot./Ctx. are task-balanced means over the groups of Table tab:roles, native latency.
% 2026-11-17: the Paused column moved here from supp_analysis.tex, which is no longer input; rows follow the method order.
% D4RT (CVPR 2026, Table 9) shading. With r = |Delta| / max(|Delta|) within each column,
% mix white with 50*sqrt(r)% red for drops or 50*r% green for gains.
\begin{table}[t]
\centering\footnotesize
\caption{\tablelang{\textbf{Ablation results on \bench.} SR (\%) of \ours and $\Delta$ (points) for each variant. Rot./Ctx.: rotation-sensitive and context-active tasks.}{\textbf{\bench 上的消融结果。}\ours 的 SR（\%）及各变体的变化（$\Delta$，百分点；红色为下降，绿色为提升）。Rot./Ctx.：旋转敏感组与环境活跃组。}}
\label{tab:analysis}
\setlength{\tabcolsep}{2.2pt}
% Equal-width numeric columns retain padding at the right edge for the cell colors.
\begin{tabularx}{\linewidth}{@{}l*{5}{Y}}
\toprule
 & \multicolumn{3}{c}{\tablelang{Success rate, all tasks}{SR，全部任务}} & \multicolumn{2}{c}{\tablelang{Groups (Native)}{任务组（原生）}}\\
\cmidrule(lr){2-4}\cmidrule(l){5-6}
 & \tablelang{Paused}{暂停} & \tablelang{Native}{原生} & +150\,ms & Rot. & Ctx.\\
\midrule
\textbf{\ours} & \textbf{90.3} & \textbf{82.1} & \textbf{34.1} & \textbf{80.4} & \textbf{82.7}\\
\midrule
\multicolumn{6}{@{}l}{\textit{\tablelang{Target motion history}{目标运动历史}}}\\
\quad \tablelang{w/o history}{无目标历史} & \cellcolor{red!50.0}$-$19.5 & \cellcolor{red!50.0}$-$19.6 & \cellcolor{green!2.9}+0.2 & \cellcolor{red!50.0}$-$24.0 & \cellcolor{red!35.4}$-$2.3\\
\multicolumn{6}{@{}l}{\textit{\tablelang{Precision allocation}{精度分配}}}\\
\quad \tablelang{Uniform coarse flow}{统一粗粒度流} & \cellcolor{green!11.3}+4.4 & \cellcolor{red!24.0}$-$4.5 & \cellcolor{red!19.2}$-$0.5 & \cellcolor{red!29.9}$-$8.6 & \cellcolor{red!50.0}$-$4.6\\
\quad \tablelang{Uniform fine flow}{统一精细流} & \cellcolor{green!3.3}+1.3 & \cellcolor{green!2.0}+0.8 & \cellcolor{red!25.7}$-$0.9 & \cellcolor{red!18.8}$-$3.4 & \cellcolor{red!37.6}$-$2.6\\
\quad \tablelang{w/o flow}{无场景流} & \cellcolor{green!7.9}+3.1 & \cellcolor{red!23.1}$-$4.2 & \cellcolor{red!32.1}$-$1.4 & \cellcolor{red!34.9}$-$11.7 & \cellcolor{red!46.0}$-$3.9\\
\multicolumn{6}{@{}l}{\textit{\tablelang{Future supervision}{未来监督}}}\\
\quad \tablelang{w/o future}{无未来监督} & \cellcolor{green!0.5}+0.2 & \cellcolor{red!23.4}$-$4.3 & \cellcolor{red!50.0}$-$3.4 & \cellcolor{red!36.2}$-$12.6 & \cellcolor{red!33.8}$-$2.1\\
\quad \tablelang{w/o $\mathcal{L}_{\text{aux}}$}{无 $\mathcal{L}_{\text{aux}}$} & \cellcolor{red!22.1}$-$3.8 & \cellcolor{red!18.2}$-$2.6 & \cellcolor{red!37.4}$-$1.9 & \cellcolor{red!34.8}$-$11.6 & \cellcolor{green!1.1}+0.1\\
\bottomrule
\end{tabularx}
\end{table}

\paragraph{Effect of target motion history.}
The pose history is the most important input. Removing it while keeping the
current pose reduces SR by 19.6 points at native latency and by 24.0 points on
rotation-sensitive tasks. The drop is the same with the world paused (19.5 points). A
single pose shows where the target is but not how it moves. The history supplies the
velocity and acceleration that commands must anticipate.

\paragraph{Effect of role-aware prediction.}
Each role needs its own form of prediction. Replacing the SE(3) trajectory with flow at
the density of the surroundings reduces SR by 4.5 points, and by 8.6 points on
rotation-sensitive tasks. At this density, the few patches on the target cannot
determine its rotation. A $16\times$ denser flow recovers the overall SR but still
trails \ours by 3.4 and 2.6 points on rotation-sensitive and context-active tasks.
Removing the sparse flow reduces SR by 4.2 points, and by 11.7 points on
rotation-sensitive tasks, because the flow shapes the patch tokens that the action
stream reads near contact. One target trajectory and 196 flow points thus match
3{,}136 uniform flow points overall and surpass them where rotation and context matter.

\paragraph{Effect of future supervision.}
Supervising only the current target state instead of its future reduces SR by 4.3
points, and by 12.6 points on rotation-sensitive tasks, although the prediction heads
never run at deployment. Prediction thus helps through the representation it
shapes. The appearance losses $\mathcal L_{\rm aux}$ also help,
mainly on rotation-sensitive tasks.

\paragraph{Prediction in a moving world.}
The three variants that weaken future prediction (\emph{uniform coarse flow},
\emph{w/o flow}, \emph{w/o future}) each lose more than 4 points at native latency
but none with the world paused. With the world paused, a policy replans every 50\,ms
and executes only the command for the capture time. At native latency, the
15.1\,ms inference of \ours delays execution to the next control boundary, 50\,ms after
capture. The representation must then encode where the target will be. Removing the history or
$\mathcal L_{\rm aux}$ instead lowers SR in both settings. Paused evaluation hides the
value of prediction, and \bench reveals it.

\subsection{Real-Robot Evaluation}
\label{sec:real}

\begin{figure}[!ht]
\centering
\includegraphics[width=\columnwidth]{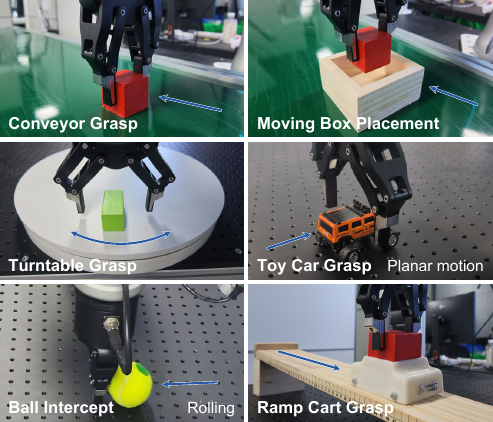}
\caption{\textbf{Real-robot tasks.} Arrows show the motion of the target or the moving
box before contact.}
\label{fig:real_trials}
\end{figure}

We evaluate on six dynamic tasks with \cfgRobot{} (Fig.~\ref{fig:real_trials}). All methods are trained on the same 600 teleoperated
demonstrations and evaluated for 20 trials per task. \ours starts from a
DROID-pretrained base \cite{droid} and uses RGB-D and online target pose estimates from
one fixed camera. ACT and $\pi_{0.5}$ use RGB images from the fixed and wrist cameras. Further
details on the tasks and platform are in the supplementary material.

% Transcribed from the author-provided six-task real-robot results. All pools 120 trials per method.
\begin{table}[t]
\centering\footnotesize
\caption{\tablelang{\textbf{Real-robot results.} SR (\%) over 20 trials per task.}{\textbf{真机结果。}每项任务试验 20 次，报告成功率（\%）。}}
\label{tab:real}
\setlength{\tabcolsep}{1.2pt}
\begin{tabular}{@{}l*{7}{c}}
\toprule
\tablelang{Method}{方法}
& \makecell{\tablelang{Conveyor}{传送带}\\\tablelang{grasp}{抓取}}
& \makecell{\tablelang{Box}{移动盒}\\\tablelang{place}{放置}}
& \makecell{\tablelang{Turntable}{转盘}\\\tablelang{grasp}{抓取}}
& \makecell{\tablelang{Toy car}{玩具车}\\\tablelang{grasp}{抓取}}
& \makecell{\tablelang{Ball}{网球}\\\tablelang{intercept}{拦截}}
& \makecell{\tablelang{Ramp cart}{斜坡小车}\\\tablelang{grasp}{抓取}}
& \tablelang{All}{总体} $\uparrow$ \\
\midrule
ACT & 0 & 0 & 0 & 0 & 10 & 0 & 1.7 \\
$\pi_{0.5}$ & 45 & 75 & 30 & 20 & 15 & 0 & 30.8 \\
\makecell[l]{\oursname\\\tablelang{(Ours)}{（本文）}} & \bestcell{70} & \bestcell{80} & \bestcell{60} & \bestcell{35} & \bestcell{35} & \bestcell{10} & \bestcell{48.3} \\
\bottomrule
\end{tabular}
\end{table}

On the robot, the scene keeps moving during inference, as in \emph{Native}. \ours
achieves the highest SR on all six tasks and 48.3\% overall, 17.5 points above
$\pi_{0.5}$ (Table~\ref{tab:real}). The largest gains, 30 and 25 points, are on
turntable and conveyor grasping. ACT, the fastest policy, succeeds in only 2 of 120
trials. On ramp cart grasping, the hardest task, only \ours succeeds
(2 of 20 trials).

\section{Conclusion}
\label{sec:conclusion}

We presented \ours, a native 4D world--action model that allocates prediction precision
by role, with precise prediction for the target, coarse prediction for its surroundings,
and none for the static scene. This allocation alleviates the tension between accurate
prediction and timely action. \ours acts in 15.1\,ms and outperforms the strongest
baselines on \bench, DOMINO, and real robots. We also introduced \bench, which evaluates
every policy at its native latency and shows that paused evaluation overestimates slow
policies. \ours currently assumes a rigid target whose pose can be tracked online, and
we leave articulated and deformable targets to future work.

\label{main:body-end}
\ifdefined\arxivversion\else\clearpage\fi
\section*{LLM Usage Statement}
\label{sec:llm_usage}
We used OpenAI Codex to assist with literature review, code implementation and
debugging, experiment execution, and language editing of the manuscript. Codex
also assisted in developing plotting code and refining figure and manuscript
layouts. The authors reviewed AI-assisted content, checked references against
original sources, and validated the implementation through testing and experiments.
The authors take full responsibility for the final manuscript, experimental
results, and conclusions.

{\small
\bibliographystyle{ieeenat_fullname}
\bibliography{references}
}
\ifdefined\withappendix
\clearpage
\setcounter{table}{0}
\setcounter{figure}{0}
\setcounter{equation}{0}
\setcounter{algorithm}{0}
\renewcommand{\thetable}{S\arabic{table}}
\renewcommand{\thefigure}{S\arabic{figure}}
\renewcommand{\theequation}{S\arabic{equation}}
\renewcommand{\thealgorithm}{S\arabic{algorithm}}
\renewcommand{\theHequation}{supp.\arabic{equation}}
\renewcommand{\theHtable}{supp.\arabic{table}}
\renewcommand{\theHfigure}{supp.\arabic{figure}}
\appendix
% Shared appendix assembly for the standalone supplement and the full English edition.
% A--D: method, benchmark, simulation results, and real-robot experiments.
% Keep supplementary float settings local to the appendix, leaving the paper unchanged.
\raggedbottom
\setcounter{topnumber}{4}
\setcounter{dbltopnumber}{4}
\setcounter{totalnumber}{6}
\renewcommand{\topfraction}{0.9}
\renewcommand{\dbltopfraction}{0.9}
\renewcommand{\textfraction}{0.08}
\renewcommand{\floatpagefraction}{0.7}
\renewcommand{\dblfloatpagefraction}{0.7}
\setlength{\textfloatsep}{12pt plus 2pt minus 2pt}
\setlength{\dbltextfloatsep}{12pt plus 2pt minus 2pt}
\setlength{\floatsep}{10pt plus 2pt minus 2pt}
\setlength{\intextsep}{8pt plus 2pt minus 2pt}
\setlength{\dblfloatsep}{10pt plus 2pt minus 2pt}
\makeatletter
\setlength{\@fptop}{0pt}
\setlength{\@fpsep}{12pt plus 2pt minus 2pt}
\setlength{\@fpbot}{0pt plus 1fil}
\setlength{\@dblfptop}{0pt}
\setlength{\@dblfpsep}{12pt plus 2pt minus 2pt}
\setlength{\@dblfpbot}{0pt plus 1fil}
\makeatother
% Appendix A (v5, 2026-11-19): method and training details of the simulated model.
% A.1 is new (rewritten from the 2026-10-24 Error Decomposition); A.2-A.3 reorganize the
% previous A.1. DROID pretraining and fine-tuning move to Appendix D.3; Figure Details removed.
% Forward references resolved by Appendix B: tab:roles (B.1), app:sweeps (B.3).
\section{Method and Training Details}
\label{app:implementation}

\subsection{Contact Error Analysis}
\label{app:decomposition}
Sec.~\mainref{sec:introduction} argues that prediction requirements differ by role.
This section derives the allocation from the error at contact. We bound the error
between the end effector and the target at the time a command executes, and assign
each term to the target, the context, or the static scene
(Sec.~\mainref{sec:formulation}). Three conclusions follow. Target errors enter the
contact directly and grow with the time between capture and execution, so the target
requires a precise forecast and a short latency. Context errors change the outcome only
when they exceed the clearance, so a coarse forecast suffices. The static scene
contributes no error that grows over time, so the current frame suffices.
Table~\ref{tab:decomposition} summarizes this correspondence.

A command of query $t$ executes at time $\tau_t+\ell$, where $\tau_t$ is the capture
time of the frame. We call $\ell$ the lead of the command. A command can execute only
after inference ends, so the lead is at least the inference latency and grows with it.
Hats denote the estimates behind a policy's commands.

\paragraph{Target position.}
The target position at execution is $\vect p(\tau_t+\ell)=\vect p(\tau_t)+\vect v\ell
+\boldsymbol\rho$, where $\vect v$ is the velocity at capture and $\boldsymbol\rho$ is
the remaining motion. Its estimate is $\hat{\vect p}(\tau_t)+\hat{\vect v}\hat\ell
+\hat{\boldsymbol\rho}$. Adding and subtracting $\vect v\hat\ell$ bounds the position
error by
\begin{equation}
e_p\le e_{\rm loc}+e_v\hat\ell+\|\vect v\|\,e_\ell+e_{\rm dyn},
\label{eq:budget_translation}
\end{equation}
with $e_{\rm loc}=\|\hat{\vect p}(\tau_t)-\vect p(\tau_t)\|$,
$e_v=\|\hat{\vect v}-\vect v\|$, $e_\ell=|\hat\ell-\ell|$, and
$e_{\rm dyn}=\|\hat{\boldsymbol\rho}-\boldsymbol\rho\|$. Each term requires different
information. The localization error $e_{\rm loc}$ depends only on the current pose. The
velocity term $e_v\hat\ell$ requires the recent motion of the target, which a single
frame does not show. The dynamics term $e_{\rm dyn}$ requires knowing how this motion
will change. Under a constant acceleration $\vect a$, $\boldsymbol\rho=\tfrac12\vect a
\ell^2$, and the history reveals $\vect a$ as well. Collisions, rolling contact, and
pushes depend on the context instead. The history matters
even without inference delay, because the robot needs time to reach a moving target.
Removing it therefore costs about as much with the world paused as at native latency (Table~\mainref{tab:analysis}).

\paragraph{Target orientation.}
We write the orientation at execution as $R(\tau_t+\ell)=\Exp(\boldsymbol\omega\ell+
\boldsymbol\eta)R(\tau_t)$ and define its estimate in the same way. For rotations about
a common axis, the orientation error $\theta$ satisfies
\begin{equation}
\theta\le\theta_{\rm loc}+e_\omega\hat\ell+\|\boldsymbol\omega\|\,e_\ell+e_\eta,
\label{eq:budget_rotation}
\end{equation}
where $\theta_{\rm loc}$, $e_\omega$, and $e_\eta$ correspond to the translation terms.
General rotations add only terms of second order in the angles. Orientation error
reaches the contact through a lever arm. A contact point at $\vect c$ in the target
frame, with $\|\vect c\|=r$, lies at $\vect p+R\vect c$ in the base frame. Its error is
$(\hat{\vect p}-\vect p)+(\hat RR^\top-I)R\vect c$. The operator norm of
$\hat RR^\top-I$ is $2\sin(\theta/2)$, so
\begin{equation}
e_c\le e_p+2r\sin(\theta/2)\le e_p+r\theta .
\label{eq:budget_contact}
\end{equation}
An error of $5^\circ$ at $r=10$\,cm moves the contact by 8.7\,mm. The target therefore
needs an explicit rigid pose. Points sampled on the target determine its rotation only
if they span this lever arm, and the few patches on a small target do not.

\paragraph{Context.}
The context affects contact in two ways. It can block the path of the robot, and it can
drive the target. Suppose the executed path keeps the robot and the carried object at
least $m$ away from the predicted positions of the context. A collision then requires a
context point to deviate from its prediction by at least $m$, so smaller errors do not
change the outcome. When the clearance $m$ is much larger than the contact tolerance
$\varepsilon$, a coarse forecast suffices for the context, whereas the target must still
be predicted within $\varepsilon$. A finer context forecast becomes necessary only when
the clearance approaches the error of the coarse one. A context that drives the target,
such as a conveyor or a ramp, enters through $e_{\rm dyn}$ instead. Its effect appears in
the motion of the target, so the future trajectory of the target already carries it.

\paragraph{Static scene.}
The static scene does not move. The current frame gives its geometry up to sensing
error, and none of its terms grows with the lead. It needs no forecast.

\paragraph{Latency.}
Every target term except $e_{\rm loc}$ and $\theta_{\rm loc}$ grows with the lead. The
velocity terms grow linearly, and $e_{\rm dyn}$ grows faster when the target
accelerates or is struck. A shorter latency reduces all of these terms at once. Conversely,
a policy with $n$ times the latency of another has about $n$ times the lead, so it needs
an $n$ times more accurate velocity estimate to keep the same velocity term.
\ours therefore keeps both its inputs and its predictions compact. The remaining
term $\|\vect v\|e_\ell$ comes from an error in the execution time. It vanishes when each command executes at its intended time, as for every method here (Appendix~\ref{app:sweeps}).

\begin{table}[!htbp]
\centering\footnotesize
\caption{\textbf{Error terms and the parts of \ours that address them.} The terms come
from Eqs.~\eqref{eq:budget_translation}--\eqref{eq:budget_contact}. The last column gives
the ablation in Table~\mainref{tab:analysis} that removes each part.}
\label{tab:decomposition}
\begin{tabularx}{\linewidth}{@{}llXl@{}}
\toprule
Term & Role & Part of \ours & Ablation \\
\midrule
$e_{\rm loc},\ \theta_{\rm loc}$ & target & current-pose token & -- \\
$e_v\hat\ell,\ e_\omega\hat\ell$ & target & SE(3) pose history & \emph{w/o history} \\
$e_{\rm dyn},\ e_\eta$ & target & future SE(3) supervision & \emph{w/o future} \\
$r\theta$ & target & SE(3) trajectory & \emph{uniform coarse flow} \\
error below $m$ & context & sparse 3D flow & \emph{w/o flow} \\
geometry & static & current RGB-D frame & -- \\
\bottomrule
\end{tabularx}
\end{table}

\subsection{Architecture}
\label{app:tokens}
Table~\ref{tab:hyper} lists the hyperparameters of \ours.
\begin{table}[!htbp]
\centering\footnotesize
\caption{\textbf{Hyperparameters of \ours on \bench.}}
\label{tab:arch}
\label{tab:hyper}
\begin{tabularx}{\linewidth}{@{}*{2}{>{\raggedright\arraybackslash}X}@{}}
\toprule
Parameter & Value \\
\midrule
\multicolumn{2}{@{}l}{\emph{Model}} \\
Input image & $224\times224$ RGB-D, one view \\
Visual encoder & \cfgVisualEncoder{}, frozen \\
History length & \cfgHistoryLen{} steps \\
Forecast horizons & \cfgHorizons{} \\
Summary tokens & \cfgMetaTokens{} \\
Layers & \cfgLayers{} \\
Width & \cfgWidth{} \\
Heads & \cfgHeads{} \\
Feed-forward width & 4096 \\
Trainable parameters & \cfgTrainableParams{} \\
\midrule
\multicolumn{2}{@{}l}{\emph{Action}} \\
Action dimension & 7 \\
Chunk length & \cfgChunk{} \\
Euler steps & \cfgEuler{} \\
\midrule
\multicolumn{2}{@{}l}{\emph{Training}} \\
Optimizer & Fused AdamW \\
Betas & $(0.9,\,0.999)$ \\
Weight decay & $10^{-4}$ \\
Learning rate & $2\times10^{-4}$ \\
LR schedule & cosine, 5{,}174 warm-up updates \\
Batch size & \cfgBatch{} windows \\
Noise draws per query & 8 \\
Epochs & 10 \\
Reported checkpoint & \cfgUpdates{} updates \\
Gradient clipping & 0.1 \\
$\lambda_F,\lambda_P,\lambda_R,\lambda_{\rm aux}$ & 0.1 \\
Training precision & BF16 compute, FP32 parameters \\
\midrule
Training hardware & 16 NVIDIA RTX PRO 6000 GPUs \\
Training time & about 24\,h \\
\bottomrule
\end{tabularx}
\end{table}

\paragraph{Scene tokens.}
The frozen \cfgVisualEncoder{} \cite{mae} turns the RGB image into \cfgPatches{} patch
tokens. Each patch token receives a learned view embedding and two linear embeddings.
One embeds the mean valid depth of the patch. The other embeds the base-frame position
$\vect x_i$ of its center pixel, back-projected with the depth map. A patch without
valid depth at its center receives no position embedding. Only the patch tokens carry 3D
rotary embeddings, computed from $\vect x_i$. The same points
$\vect x_i$ are the source points of the sparse 3D flow.

\paragraph{State tokens.}
The \cfgHistoryLen{} target-state tokens describe the target at the offsets
$-350,-300,\ldots,0$\,ms. Each is a 25-dimensional vector
$(\vect p,\log R,\Delta\vect p,\log\Delta R,\Delta t,\vect v,\boldsymbol\omega,\vect a,
\boldsymbol\alpha)$. Slots 0--6 describe the seven intervals between consecutive
offsets, and slot 7 holds the current pose. Velocities and
accelerations are finite differences over the elapsed times. Each dimension is
standardized, and invalid dimensions are set to zero. Three
flags mark whether the pose, the interval, and the acceleration are valid. A slot with
no valid feature is masked out of attention. The robot tokens hold the end-effector
position, rotation vector, and gripper opening at the same offsets.
The \cfgMetaTokens{} summary tokens are learned embeddings.

\paragraph{Backbone and heads.}
In each layer, world and action tokens attend under the mask of
Fig.~\mainref{fig:architecture}(b). The two streams share the query, key, value, and
output projections of the attention. Each stream has its own feed-forward network with
SiLU activation and a hidden width of 4096. World tokens use layer normalization, and
action tokens use adaptive layer normalization conditioned on the flow time $u$. Each
prediction head is a linear layer on the final world-stream states, except the target
reconstruction, which is a two-layer MLP. The pose head reads the mean summary state
($\mathcal L_P$, Eq.~\mainref{eq:poseloss}). The flow head reads each patch state and
gives the displacement and visibility logit of its point at every horizon
($\mathcal L_F$, Eq.~\mainref{eq:flowloss}). All heads are skipped at deployment.

\paragraph{Inference time.}
\label{app:latency_results}
For each of the 900 profiled queries in Table~\ref{tab:supp_timing}, we subtract the three stage times from the query's wall time to obtain other query time.

\begin{table}[!htb]
\centering\footnotesize
% Transcribed from results/realtime_c/stage-timing-20261104.json (Table S3).
% First row: table1_calibration (the p95 reported as 15.1 ms in the paper).
% Full query and its four stages: profiled_total and the stage records of the same profiled run;
% other query time is the per-query residual (wall time minus the three stages), then its quantiles.
% Prediction heads: separate eager forward-only timing.
\caption{\tablelang{\textbf{\ours timing.} The benchmark and stage profile use separate runs of the compiled BF16 policy at batch size 1 on one \cfgGPU{}. Prediction heads were timed in eager BF16 on cached world features from one fixed window.}{\textbf{\ours 计时。}基准计时和分阶段计时分别进行，均在单块 \cfgGPU{} 上运行编译后的 BF16 策略，批大小为 1。预测头以 BF16 在 eager 模式下单独计时，输入为来自同一个固定观测窗口的缓存世界流特征。}}
\label{tab:supp_timing}
\setlength{\tabcolsep}{4pt}
\begin{tabularx}{\linewidth}{@{}>{\raggedright\arraybackslash}Xrr@{}}
\toprule
\tablelang{Time in ms}{耗时 / ms} & p50 & p95 \\
\midrule
\tablelang{Table~\mainref{tab:main} benchmark}{表~\mainref{tab:main} 基准计时} & 14.59 & \textbf{15.11} \\
\midrule
\tablelang{Full query with profiling}{分阶段测量的查询总耗时} & 14.66 & 15.57 \\
\quad\tablelang{Observation encoding and H2D}{观测编码与 H2D 传输} & 4.83 & 5.28 \\
\quad\tablelang{World stream}{世界流} & 3.93 & 4.22 \\
\quad\tablelang{Action stream}{动作流} & 4.27 & 4.41 \\
\quad\tablelang{Other query time}{其他查询时间} & 1.64 & 1.73 \\
\midrule
\tablelang{Prediction heads, forward only}{预测头，仅前向} & 0.37 & 0.41 \\
\bottomrule
\end{tabularx}

\end{table}

\subsection{Training}
\label{app:training}
\paragraph{Action labels.}
Each action token embeds one seven-dimensional command, which consists of a
base-frame translation increment, a rotation vector that left-multiplies the current
orientation, and a gripper command. Both increments are relative to the measured
end-effector pose when the command is applied. Command $j$ of query $t$ is labeled with
the demonstration command issued at $\tau_t+j\Delta_a$. In the post-impact grasp, labels follow the issue time (Eq.~\eqref{eq:apptime}). Commands are
standardized with training statistics.

\paragraph{Prediction labels.}
Flow labels come from the simulator. The surface point at each patch center
follows the ground-truth rigid motion of its MuJoCo geometry. A future point is visible
if it projects onto the same geometry within a depth tolerance of
$\max(1\,\mathrm{mm},\,0.001z)$, where $z$ is its depth. Invisible points still supervise displacement, with visibility label zero. Displacements are divided
by the scale $s=0.715$\,m, and the visibility term has weight 1. Pose labels are the target's ground-truth poses. Horizons beyond the last frame of a demonstration are
excluded from both losses.

\paragraph{Auxiliary losses.}
$\mathcal L_{\rm aux}$ sums four terms. Three of them are squared errors on future
metric depth at valid pixels, on the future RGB of surfaces that move by more than
1\,mm, and on future layer-normalized DINOv3 features \cite{dinov3}.
The fourth reconstructs the RGB of the current target region from the mean summary
state with an absolute error.

\paragraph{Optimization.}
Each batch holds 512 windows. A window contains up to eight consecutive queries from
one demonstration, and each query has its own causal history. The world stream runs
once per query, as at inference. Its keys and values are reused for eight independent
draws of action noise and flow time, and the action loss averages over these draws.
All experiments, including the baselines, the ablations, and DROID pretraining, ran on
64 NVIDIA RTX PRO 6000 and 32 NVIDIA H100 GPUs. One training run of \ours takes about
24\,h on 16 of these RTX PRO 6000 GPUs.

% Appendix B (v5.1a, 2026-11-20): v5.1 with the corrections required by docs/writing/appendix_B_verification.md.
% Structure mirrors Sec. 4 of the main paper: tasks, construction (4.1), evaluation protocol (4.2).
\section{\bench and Evaluation Protocol}
\label{app:benchmark}

\subsection{Tasks}
\label{app:tasks}
\label{app:roles}
Table~\ref{tab:tasks} lists the target and context of each task (Sec.~\mainref{sec:formulation}).

The tasks differ in which terms of the decomposition in
Appendix~\ref{app:decomposition} matter. The target moves in every task, so the velocity
term always applies. In the seven rotation-sensitive tasks, the target also rotates, and
its orientation matters at contact. In the nine context-active tasks, the context
limits the clearance or drives the target. In the oscillating, capture, and impact tasks, the motion of the target changes within the episode through oscillation, gravity, or a collision, so its current velocity is not enough. Fig.~\mainref{fig:bench_overview}c shows all tasks.

% Protocol table (v5.1, 2026-11-20): rows grouped by family: the fourteen OnTimeBench tasks with target, context, motion, and success.
% Context follows Sec. 3.1: obstacles the robot must clear and drivers that move the target; static bodies are not listed.
% Context and Motion checked in docs/writing/appendix_B_verification.md.
\begin{table*}[!htbp]
\centering\footnotesize
\caption{\tablelang{\textbf{The fourteen \bench tasks.}
The target is the body whose pose the policy receives. For placement, it is the receptacle. For wire passage, it is the wire entry, and the rest of the wire is context.
The context contains the obstacles and drivers of each task, and a dash means none.
Speed is the linear speed of the target in m/s, or the angular speed of the wire. For the capture tasks, it is the speed at which the target leaves the ramp, measured over 100 held-out reference trajectories.
A grasp requires contact on both fingerpads, a lift of at least 0.05\,m, and a continuous hold of 0.5\,s.
An episode fails on unsafe contact, on leaving the workspace, on a drop, or at the time limit.
Rot. marks the rotation-sensitive tasks, in which the target rotates before contact and
its orientation matters at contact. Ctx. marks the context-active tasks, which have a
context. Table~\mainref{tab:analysis} reports both groups.}{\textbf{\bench 的十四项任务。}
目标是策略接收其位姿的物体。放置任务的目标是容器。导线穿越任务的目标是导线入口，导线其余部分属于环境。
环境包括各任务中的障碍物和驱动物，破折号表示没有环境物体。
速度为目标的线速度（m/s），或导线的角速度。捕获任务的速度为目标离开斜坡时的速度，在 100 条留出参考轨迹上测得。
抓取要求双侧指垫接触、抬升至少 0.05\,m，并连续保持 0.5\,s。
发生不安全接触、离开工作空间、掉落或达到时限时，回合失败。
Rot. 表示旋转敏感任务，即目标在接触前旋转、且朝向影响接触的任务。Ctx. 表示环境活跃任务，即有环境物体的任务。表~\mainref{tab:analysis} 报告两组的结果。}}
\label{tab:tasks}
\label{tab:roles}
\setlength{\tabcolsep}{3.75pt}
\begin{tabular*}{\textwidth}{@{\extracolsep{\fill}}llllclcc@{}}
\toprule
\tablelang{Task}{任务} & \tablelang{Target}{目标} & \tablelang{Context}{环境} & \tablelang{Motion}{运动} & \tablelang{Speed}{速度} & \tablelang{Success}{成功条件} & \tablelang{Limit (s)}{时限（s）} & \tablelang{Group}{任务组} \\
\midrule
\multicolumn{8}{@{}l}{\textit{\tablelang{Grasp}{抓取}}} \\
\tablelang{Harmonic grasp}{谐波抓取} & \tablelang{block}{物块} & -- & \tablelang{harmonic}{谐波} & 0.03--0.20 & \tablelang{grasp}{抓取} & 8 & Rot. \\
\tablelang{Conveyor grasp}{传送带抓取} & \tablelang{part}{零件} & \tablelang{belt}{传送带} & \tablelang{belt}{传送带} & 0.04--0.10 & \tablelang{grasp}{抓取} & 8 & Ctx. \\
\tablelang{Translating grasp}{平移抓取} & \tablelang{part}{零件} & -- & \tablelang{oscillating}{振荡} & 0.04--0.15 & \tablelang{grasp}{抓取} & 8 & -- \\
\tablelang{Rotating grasp}{旋转抓取} & \tablelang{part}{零件} & -- & \tablelang{oscillating, yaw}{振荡、偏航} & 0.04--0.15 & \tablelang{grasp}{抓取} & 8 & Rot. \\
\tablelang{Occluded grasp}{遮挡抓取} & \tablelang{part}{零件} & \tablelang{belt, canopy}{传送带、遮挡罩} & \tablelang{belt}{传送带} & 0.04--0.10 & \tablelang{grasp without canopy contact}{抓取，不接触遮挡罩} & 8 & Ctx. \\
\midrule
\multicolumn{8}{@{}l}{\textit{\tablelang{Place}{放置}}} \\
\tablelang{Moving receptacle}{移动容器} & \tablelang{receptacle}{容器} & -- & \tablelang{linear}{直线} & 0.03--0.06 & \tablelang{held object at rest inside}{所持物体静止于容器内} & 10 & -- \\
\tablelang{Rotating receptacle}{旋转容器} & \tablelang{T-shaped receptacle}{T 形容器} & -- & \tablelang{linear and yaw}{直线与偏航} & 0.03--0.06 & \tablelang{held T-part at rest inside}{所持 T 形件静止于容器内} & 10 & Rot. \\
\midrule
\multicolumn{8}{@{}l}{\textit{\tablelang{Pass}{穿越}}} \\
\tablelang{Wire passage}{导线穿越} & \tablelang{wire entry}{导线入口} & \tablelang{wire}{导线} & \tablelang{rotation}{旋转} & 2.5--7.5$^\circ$/s & \tablelang{align within 30\,mm, pass without contact}{30\,mm 内对准，无接触穿过} & 21 & Rot., Ctx. \\
\tablelang{Full wire passage}{完整导线穿越} & \tablelang{wire entry}{导线入口} & \tablelang{wire}{导线} & \tablelang{rotation}{旋转} & 2.5--7.5$^\circ$/s & \tablelang{grasp wand, pass, exit, release}{抓杆、穿过、离开、释放} & 32 & Rot., Ctx. \\
\midrule
\multicolumn{8}{@{}l}{\textit{\tablelang{Capture}{捕获}}} \\
\tablelang{Ball capture}{小球捕获} & \tablelang{ball}{小球} & \tablelang{ramp}{斜坡} & \tablelang{rolling}{滚动} & 0.67--1.14 & \tablelang{grasp before the tray}{触及托盘前抓取} & 6 & Ctx. \\
\tablelang{Ball capture on table}{桌面小球捕获} & \tablelang{ball}{小球} & \tablelang{ramp}{斜坡} & \tablelang{rolling}{滚动} & 0.59--1.03 & \tablelang{grasp while rolling}{滚动中抓取} & 6 & Ctx. \\
\tablelang{Slider capture}{滑块捕获} & \tablelang{slider}{滑块} & \tablelang{ramp}{斜坡} & \tablelang{sliding}{滑动} & 0.55--1.12 & \tablelang{grasp while sliding}{滑动中抓取} & 6 & Ctx. \\
\tablelang{Can capture}{罐体捕获} & \tablelang{can}{罐体} & \tablelang{ramp}{斜坡} & \tablelang{rolling}{滚动} & 0.59--0.97 & \tablelang{grasp before tipping}{倾倒前抓取} & 6 & Rot., Ctx. \\
\midrule
\multicolumn{8}{@{}l}{\textit{\tablelang{Impact}{撞击}}} \\
\tablelang{Post-impact grasp}{撞击后抓取} & \tablelang{part}{零件} & \tablelang{belt, striker}{传送带、撞击器} & \tablelang{impact}{撞击} & 0.12--0.14 & \tablelang{grasp along its axis}{沿抓取轴抓取} & 6 & Rot., Ctx. \\
\bottomrule
\end{tabular*}
\end{table*}

\subsection{Construction}
\label{app:splits}
\paragraph{Simulation.}
The tasks run in MuJoCo 3.11.0 and robosuite 1.5.2 \cite{mujoco,robosuite} with a
Franka Panda arm and a parallel gripper. The operational-space controller of robosuite
tracks the commanded end-effector pose. The control period is 50\,ms, and the physics
step is 2\,ms. A fixed third-person camera with a $40^\circ$ vertical field of view
renders $224\times224$ RGB-D images. Actions are the seven-dimensional commands of
Appendix~\ref{app:training}. Each translation component is clipped to 0.10\,m, and each
rotation component to 0.5\,rad.

\paragraph{Scenes.}
Each scene samples the target's initial state and motion parameters uniformly
within task-specific ranges. Table~\ref{tab:tasks} gives the speed ranges. Other sampled
parameters include harmonic amplitudes of up to 35\,mm at 0.25--0.7\,Hz, yaw rates of up
to 1.2\,rad/s, and ramp slopes of $5^\circ$, $10^\circ$, or $15^\circ$. In the occluded grasp, the target is
hidden for 0.5--3.0\,s and then reappears. Only physically feasible scenes are kept, and every method replays the same stored scenes.

\paragraph{Demonstrations.}
The expert is a task-specific closed-loop state machine. It reads the privileged
simulator state and anticipates the motion of the target, for example by predicting the
interception point in the capture tasks. A demonstration is kept if the expert succeeds,
if a replay of its actions reproduces the episode, and if its path and jerk pass quality
checks. The training set holds 800 demonstrations per task, 11{,}200 in total. They are
recorded at 20\,Hz with RGB-D images, robot states, commands, and simulator ground
truth.

\paragraph{Splits.}
We select configurations on 1{,}400 development scenes, 100 per task. All results come from 1{,}400 held-out test scenes, also 100 per task, sampled from the same ranges with seeds unused in training and development.

\subsection{Evaluation Protocol}
\label{app:sweeps}
% Queue the baseline and per-task tables before the next two-column page.
% Protocol table: Dynamic-14 baseline adaptations from BASE1 and the v3 endpoint records.
% PUMA uses the resolved 40k training configuration and parameter summary in
% BASE1/results/puma_dynamic14_20261101/h100/training-final.zip.
\tablelang{\begin{table}[!tbp]}{\begin{table}[!htbp]}
\centering\footnotesize
\caption{\tablelang{\textbf{Baseline training on \bench.} Updates$\times$batch gives the optimizer updates and the global batch size.
$H$ is the number of commands returned per query. The parameter counts include trainable parameters only.
}{\textbf{\bench 上的基线训练。}「更新数$\times$batch」一列给出优化器更新次数和全局 batch 大小。
$H$ 为每次查询返回的指令数。参数量只计入可训练参数。}}
\label{tab:baselines}
\setlength{\tabcolsep}{2.6pt}
\begin{tabular*}{\linewidth}{@{\extracolsep{\fill}}llccc@{}}
\toprule
\tablelang{Method}{方法} & \tablelang{Initialization}{初始化} & \tablelang{Updates$\times$batch}{更新数$\times$batch} & $H$ & \tablelang{Params}{参数量} \\
\midrule
ACT \cite{act} & \tablelang{scratch}{从头训练} & $50{,}000\times32$ & 100 & 83.9M \\
DP \cite{diffusionpolicy} & \tablelang{scratch}{从头训练} & $50{,}000\times32$ & 15 & 288.7M \\
$\pi_{0.5}$ \cite{pi05} & \texttt{pi05\_base} & $30{,}000\times256$ & 10 & 3,353.4M \\
$\pi_{0.5}$+SE(3) & \texttt{pi05\_base} & $30{,}000\times256$ & 10 & 3,357.7M \\
DynamicVLA \cite{dynamicvla} & \tablelang{official}{官方} & $50{,}000\times44$ & 20 & 99.1M \\
PUMA \cite{domino} & \tablelang{official}{官方} & $40{,}000\times64$ & 16 & 4,516.5M \\
OpenWAM-$\alpha$ \cite{openwam_alpha} & \tablelang{official}{官方} & $34{,}490\times192$ & 32 & 6,021.2M \\
DynamicWAM \cite{dynamicwam} & \tablelang{official}{官方} & $40{,}000\times128$ & 16 & 988.8M \\
DynamicWAM+SE(3) & \tablelang{official}{官方} & $40{,}000\times128$ & 16 & 989.4M \\
\midrule
\oursentry & \tablelang{scratch}{从头训练} & $\cfgUpdates\times512$ & \cfgChunk & \cfgTrainableParams \\
\bottomrule
\end{tabular*}
\end{table}

% Generated by scripts/build_realtime_per_task.py from results/realtime_c/native-per-task-20261104.json.
% Native real-time: measured method-specific inference latency + 0 ms extra; one training seed.
\tablelang{\begin{table*}[!t]}{\begin{table}[H]}
\centering\footnotesize
\caption{\tablelang{\textbf{Per-task SR} (\%) on \bench at native latency. Each method is evaluated on the same 100 held-out scenes per task. The last two rows give the first-contact position error $e_p$ (mm), averaged over the 13 tasks with a contact, and the number of episodes with a contact out of 1,400. A dash marks a method that makes no contact on some task. DP: Diffusion Policy. DVLA: DynamicVLA. OWAM: OpenWAM-$\alpha$. OW-off: OpenWAM-$\alpha$ without cache and compilation. DWAM: DynamicWAM. +: with our SE(3) history tokens. Bold and light blue mark the best and second-best values.}
{\textbf{\bench 上的逐任务成功率}（\%），原生延迟。每个方法都在相同的 100 个留出场景上评测每个任务。最后两行是首次接触位置误差 $e_p$（mm，在 13 个有接触的任务上平均）和 1,400 个回合中发生接触的回合数；破折号表示该方法在某个任务上没有接触。DP：Diffusion Policy。DVLA：DynamicVLA。OWAM：OpenWAM-$\alpha$。OW-off：关闭缓存与编译的 OpenWAM-$\alpha$。DWAM：DynamicWAM。+：加入本文的 SE(3) 历史 token。粗体和浅蓝色分别标出最佳和次佳。}}
\label{tab:supp_per_task}
\setlength{\tabcolsep}{3pt}
\begin{tabularx}{\linewidth}{@{}l*{10}{Y}c}
\toprule
\tablelang{Task}{任务} & ACT & DP & $\pi_{0.5}$ & $\pi_{0.5}$+ & DVLA & PUMA & OWAM & OW-off & DWAM & DWAM+ & \makecell{\oursname\\\tablelang{(Ours)}{（本文）}} \\
\midrule
\tablelang{Harmonic grasp}{谐波抓取} & 49 & 31 & 51 & 84 & 39 & \secondcell{85} & 41 & 1 & 50 & 47 & \bestcell{99} \\
\tablelang{Conveyor grasp}{传送带抓取} & 45 & 45 & \secondcell{96} & 62 & 89 & 89 & 62 & 0 & 59 & 60 & \bestcell{100} \\
\tablelang{Translating grasp}{平移抓取} & 14 & 33 & 59 & 42 & \secondcell{61} & 51 & 53 & 0 & 53 & 18 & \bestcell{77} \\
\tablelang{Rotating grasp}{旋转抓取} & 24 & 20 & 32 & 53 & 26 & \secondcell{67} & 42 & 3 & 30 & 35 & \bestcell{71} \\
\tablelang{Occluded grasp}{遮挡抓取} & 30 & 28 & 37 & 60 & 80 & \secondcell{84} & 34 & 0 & 44 & 38 & \bestcell{100} \\
\midrule
\tablelang{Moving receptacle}{移动容器} & 26 & 24 & 34 & \secondcell{63} & 41 & 24 & 27 & 0 & 8 & 6 & \bestcell{93} \\
\tablelang{Rotating receptacle}{旋转容器} & 0 & 0 & 1 & \secondcell{11} & 1 & 0 & 0 & 0 & 0 & 0 & \bestcell{66} \\
\midrule
\tablelang{Wire passage}{导线穿越} & \secondcell{99} & \secondcell{99} & \bestcell{100} & 89 & \secondcell{99} & 87 & 50 & 58 & 23 & 78 & \bestcell{100} \\
\tablelang{Full wire passage}{完整导线穿越} & 0 & 3 & 1 & 0 & \secondcell{12} & 0 & 0 & 0 & 0 & 0 & \bestcell{40} \\
\midrule
\tablelang{Ball capture}{小球捕获} & 36 & 16 & \bestcell{46} & 29 & 41 & 39 & \secondcell{45} & 13 & 24 & 20 & 38 \\
\tablelang{Ball capture on table}{桌面小球捕获} & 71 & 81 & \bestcell{100} & \bestcell{100} & 88 & \bestcell{100} & \secondcell{99} & 88 & 98 & \bestcell{100} & \bestcell{100} \\
\tablelang{Slider capture}{滑块捕获} & 56 & 30 & 58 & 62 & 28 & \secondcell{76} & 71 & 31 & 47 & 63 & \bestcell{79} \\
\tablelang{Can capture}{罐体捕获} & 91 & 31 & 98 & \secondcell{99} & 72 & \bestcell{100} & \bestcell{100} & 5 & 96 & \bestcell{100} & \bestcell{100} \\
\midrule
\tablelang{Post-impact grasp}{撞击后抓取} & 42 & 5 & 49 & \secondcell{68} & 39 & 57 & 21 & 22 & 43 & 19 & \bestcell{87} \\
\midrule
\tablelang{Task-balanced}{任务等权} & 41.6 & 31.9 & 54.4 & 58.7 & 51.1 & \secondcell{61.4} & 46.1 & 15.8 & 41.1 & 41.7 & \bestcell{82.1} \\
\midrule
$e_p$ (mm) & 32.2 & 36.1 & 31.3 & 34.8 & 30.0 & \secondcell{27.2} & 31.8 & -- & 31.2 & 36.1 & \bestcell{22.0} \\
\tablelang{Contacts}{接触回合数} & 1110 & 946 & 1237 & 1191 & 1204 & 1272 & 1115 & 493 & 1193 & 1029 & 1285 \\
\bottomrule
\end{tabularx}
\tablelang{\end{table*}}{\end{table}}

% Generated by scripts/build_domino_per_task.py from results/domino_baselines/per-task-20261122.csv.
\tablelang{\begin{table*}[!t]}{\begin{table}[H]}
\centering\footnotesize
\caption{\tablelang{\textbf{Per-task SR} (\%) \textbf{and MS} on DOMINO over 100 scenes per task. MS: manipulation score. Bold marks the best value.}{\textbf{DOMINO 上的逐任务成功率}（\%）\textbf{和操作得分}，每项任务 100 个场景。MS：操作得分。粗体标出最佳值。}}
\label{tab:supp_domino_per_task}
\setlength{\tabcolsep}{3pt}
\renewcommand{\arraystretch}{0.95}
\begin{tabular*}{\linewidth}{@{\extracolsep{\fill}}l*{8}{c}}
\toprule
 & \multicolumn{2}{c}{$\pi_{0.5}$} & \multicolumn{2}{c}{DynamicWAM} & \multicolumn{2}{c}{PUMA} & \multicolumn{2}{c}{\oursname{} \tablelang{(Ours)}{（本文）}} \\
\cmidrule(lr){2-3}\cmidrule(lr){4-5}\cmidrule(lr){6-7}\cmidrule(lr){8-9}
\tablelang{Task}{任务} & SR & MS & SR & MS & SR & MS & SR & MS \\
\midrule
\multicolumn{9}{@{}l}{\emph{\tablelang{Intercept}{拦截}}} \\
Adjust Bottle & 20 & 39.8 & 19 & 34.3 & \bestcell{68} & \bestcell{80.1} & 37 & 57.7 \\
Move Can Pot & 18 & 44.0 & 7 & 40.0 & 28 & 51.0 & \bestcell{52} & \bestcell{67.7} \\
Move Playingcard Away & 7 & 42.9 & 9 & 36.8 & \bestcell{18} & 51.4 & \bestcell{18} & \bestcell{59.1} \\
Place A2B Left & 0 & 21.7 & 4 & 32.6 & 11 & 41.4 & \bestcell{17} & \bestcell{47.8} \\
Place A2B Right & 2 & 26.6 & 1 & 35.6 & 3 & 34.7 & \bestcell{19} & \bestcell{46.9} \\
Rotate QRcode & 11 & 32.9 & 4 & 25.0 & \bestcell{19} & 40.1 & 18 & \bestcell{53.1} \\
Shake Bottle & 68 & 73.9 & 56 & 64.1 & \bestcell{76} & \bestcell{82.4} & 41 & 66.5 \\
Shake Bottle Horiz. & 64 & 69.5 & 34 & 45.8 & \bestcell{67} & \bestcell{76.3} & 33 & 61.5 \\
\midrule
\multicolumn{9}{@{}l}{\emph{\tablelang{Track}{跟踪}}} \\
Beat Block Hammer & 0 & 12.8 & 30 & 42.4 & 16 & 29.9 & \bestcell{96} & \bestcell{96.6} \\
Click Alarmclock & 35 & 38.9 & 30 & 35.9 & 8 & 13.2 & \bestcell{55} & \bestcell{61.3} \\
Click Bell & 10 & 21.3 & \bestcell{35} & \bestcell{43.1} & 0 & 10.5 & 26 & 40.5 \\
Move Pillbottle Pad & 1 & 11.0 & 34 & 41.7 & 11 & 20.0 & \bestcell{52} & \bestcell{59.4} \\
Move Stapler Pad & 0 & 13.7 & 1 & 21.4 & 0 & 18.3 & \bestcell{7} & \bestcell{28.7} \\
Place Bread Basket & 2 & 13.6 & 43 & 51.0 & 17 & 27.0 & \bestcell{60} & \bestcell{67.0} \\
Place Container Plate & 8 & 17.0 & 65 & 70.4 & 62 & 67.4 & \bestcell{88} & \bestcell{90.3} \\
Place Empty Cup & 2 & 18.1 & 34 & 45.5 & 20 & 35.3 & \bestcell{94} & \bestcell{95.5} \\
Place Fan & 0 & 16.1 & 2 & 16.3 & 5 & 18.4 & \bestcell{61} & \bestcell{70.7} \\
Place Mouse Pad & 1 & 12.8 & \bestcell{27} & \bestcell{41.5} & 4 & 21.8 & 16 & 32.6 \\
Place Object Scale & 0 & 8.8 & 10 & 23.3 & 2 & 12.5 & \bestcell{21} & \bestcell{32.9} \\
Place Object Stand & 4 & 16.7 & 24 & 34.8 & 3 & 18.2 & \bestcell{70} & \bestcell{76.9} \\
Place Phone Stand & 0 & 13.3 & 17 & 25.8 & 1 & 13.0 & \bestcell{45} & \bestcell{57.0} \\
Place Shoe & 27 & 39.8 & 42 & 55.2 & 31 & 44.2 & \bestcell{47} & \bestcell{59.4} \\
Press Stapler & 33 & 41.7 & 65 & 70.4 & 2 & 15.6 & \bestcell{69} & \bestcell{74.1} \\
Stamp Seal & 7 & 19.6 & 52 & 60.7 & 13 & 27.1 & \bestcell{81} & \bestcell{84.6} \\
\midrule
\tablelang{Intercept mean}{拦截均值} & 23.8 & 43.9 & 16.8 & 39.3 & \bestcell{36.3} & 57.2 & 29.4 & \bestcell{57.6} \\
\tablelang{Track mean}{跟踪均值} & 8.1 & 19.7 & 31.9 & 42.5 & 12.2 & 24.5 & \bestcell{55.5} & \bestcell{64.2} \\
\tablelang{All}{全部} & 13.3 & 27.8 & 26.9 & 41.4 & 20.2 & 35.4 & \bestcell{46.8} & \bestcell{62.0} \\
\bottomrule
\end{tabular*}
\tablelang{\end{table*}}{\end{table}}

\paragraph{Executor.}
All methods share one executor on the 50\,ms control clock
(Algorithm~\ref{alg:executor}). Command $j$ of query $t$ is meant for the control
boundary $\tau_t+j\Delta_a$ and executes at
\begin{equation}
\tapp_{t,j}=\tau_t+j\Delta_a+d_{\rm task}.
\label{eq:apptime}
\end{equation}
In the post-impact grasp, the environment applies each command 50\,ms after it is
issued, so $d_{\rm task}=50$\,ms. In the other tasks, $d_{\rm task}=0$. A query with
latency $L_t$ hands over its chunk at the first boundary at or after its end, which lies
$k_t=\lceil L_t/\Delta_a\rceil$ periods after capture. Latency thus counts in whole
control periods, as in RTC \cite{rtc}. Execution starts from command $k_t$ and skips the commands for earlier boundaries. The next query captures its frame at this boundary. While it runs, the current chunk continues, so each policy
replans as often as its latency allows. A chunk holds $H_m$ commands, the native chunk
length of each method (Table~\ref{tab:baselines}), and is dropped if $k_t\ge H_m$. When
no chunk is active, the controller holds the current arm target and gripper command.
Inference runs asynchronously with execution, as in DynamicVLA \cite{dynamicvla}. No method uses RTC's chunk inpainting.

\begin{algorithm}[t]
\caption{\textbf{Execution at native latency.} All methods share this executor. $L_t$ is drawn from the latencies measured for each method.}
\label{alg:executor}
\footnotesize
\begin{algorithmic}[1]
\State capture an observation at $\tau_0=0$ and start query 0
\For{each control boundary $b=0,\Delta_a,2\Delta_a,\ldots$}
  \If{query $t$ has ended, that is, $\tau_t+L_t\le b$}
    \State $k_t\gets(b-\tau_t)/\Delta_a$
    \If{$k_t<H_m$}
      \State switch to chunk $t$, starting from command $k_t$
    \Else
      \State clear the active chunk
    \EndIf
    \State capture an observation at $\tau_{t+1}=b$ and start query $t+1$
  \EndIf
  \If{the active chunk has a command for $b$}
    \State issue this command, which executes at $b+d_{\rm task}$
  \Else
    \State hold the current arm target and gripper command
  \EndIf
  \State advance physics by $\Delta_a$ in steps of 2\,ms
\EndFor
\end{algorithmic}
\end{algorithm}

\paragraph{Protocols.}
In \emph{Native}, each query draws $L_t$ from the latencies of its method measured on
one \cfgGPU{}. Physics advances by this amount, however long the rollout actually takes
on the \cfgEvalGPU{}. The latency sweep adds 50, 100, 150, or
200\,ms to every $L_t$. Its 150\,ms point is the +150\,ms setting of
Table~\mainref{tab:main}. In \emph{Paused}, physics stops during each query. The policy observes at every control boundary, and its chunk starts at once from command 0.

\paragraph{Latency measurement.}
\label{app:timing}
We time each method alone on one \cfgGPU{} at batch size 1. Each baseline runs in the
precision of its release, and \ours runs in compiled BF16 with TF32 disabled. Each
method runs three rounds of 30 warm-up calls and 300 measured calls on recorded observations with their causal history. A timed query starts from the newest
observation and ends when its first command is available. We call this interval the
time-to-action. It includes all computation that needs the newest frame, namely preprocessing, image encoding, the online flow
and motion descriptors of PUMA and DynamicWAM, sampling, and action decoding. PUMA is
timed with its three official cameras and the other baselines with two, which changes
their p95 time-to-action by at most 3.8\,ms relative to the agent view.

\subsection{Metrics}
\label{app:metrics}
\label{app:statistics}
The success rate (SR) weights the 14 tasks equally. We report the 95th percentile (p95) of the time-to-action. The position error at first contact $e_p$ is the distance of the end effector from the expert's contact frame on the target. In the placement tasks, $e_p$ uses the first contact of the held object with
the receptacle and the expert's placement pose. $e_p$ averages over all tasks
except the wire passage, which succeeds without contact, and over episodes with contact. Paired 95\% intervals resample held-out scenes per task.

\subsection{Baselines}
\label{app:baselines}
\paragraph{Common conditions.}
Each baseline keeps its official architecture, chunk length, initialization, and
training recipe (Table~\ref{tab:baselines}). It trains on the same 11{,}200
demonstrations, mapped to its action space through the same Cartesian controller, and
is configured on the development scenes. Baselines are trained and evaluated on the
agent view, and any wrist-camera slot is masked. The histories follow
Table~\mainref{tab:main}. DP uses the DDIM sampler of its
real-robot release with 16 steps. OpenWAM-$\alpha$ runs with video decoding disabled and
is timed with and without its official cache and compilation.

\paragraph{SE(3)-augmented baselines.}
$\pi_{0.5}$+SE(3) and DynamicWAM+SE(3) receive the eight target-state tokens of
Appendix~\ref{app:tokens} from the same pose source and with the same validity flags. A
two-layer MLP with a zero-initialized output layer maps them to the width of each model.
In $\pi_{0.5}$, they follow the image tokens in the prefix, where the action expert
attends to them. In DynamicWAM, they join its kinematic descriptors. Training follows
the recipe of the base model.

% Appendix C (v7 layout draft, 2026-11-22): additional simulation results.
% Three subsections, in the order of Sec. 5 of the main paper.
% Fig. S6 is typeset with the full-page floats in appendix/real.tex.
% Table S6 (timing) now lives in Appendix A.2 (label app:latency_results).
% The input-sensitivity study (P4) is not part of this draft.
\section{Additional Simulation Results}
\label{app:results}

\subsection{Per-Task Results}
\label{app:per_task}
Table~\ref{tab:supp_per_task} reports the per-task SR of every method in Table~\mainref{tab:main}.

\subsection{Results on DOMINO}
\label{app:external}
We use the 24 single-arm Franka tasks of the clean level-1 setting of DOMINO
\cite{domino} with its demonstrations, scenes, and native evaluator. The evaluator pauses
physics during inference. The manipulation score (MS) scales route completion by
penalties for leaving the workspace or view and for collisions. \ours reads the ground-truth pose of the instructed target.
Table~\ref{tab:supp_domino_per_task} gives per-task results.

\paragraph{Data and training.}
The data split has 1{,}200 training and 240 validation demonstrations, and the test uses
another 2{,}400 scenes, 100 per task. Policies execute absolute joint-position targets
with DOMINO's native controller. Their state and action projections use its
eight-dimensional joint space, normalized on training data. The baselines use their official recipes. DynamicWAM and PUMA are initialized from
their official checkpoints and fine-tuned on our Franka demonstrations. These
checkpoints were trained on ALOHA-AgileX simulation demonstrations of all 35 dynamic
tasks of DOMINO. $\pi_{0.5}$ starts from the openpi \texttt{pi05\_base} checkpoint, and \ours uses no robot pretraining.

\subsection{Rollouts and Forecasts}
\label{app:prediction}
Fig.~\ref{fig:rollouts} shows rollouts of \ours on eight tasks from all five families, with the forecast at each frame.

\section{Real-Robot Experiments}
\label{app:real}
\widowpenalty=10000

\paragraph{Platform and tasks.}
\label{app:real_platform}
The robot is \cfgRobot{}. Policies send tool-center-point poses and gripper targets at 20\,Hz. Two RealSense D435 cameras observe the scene. One is fixed
and calibrated to the robot base, and the other is mounted on the wrist.
\ours uses RGB-D from the fixed camera, and the baselines use RGB from both cameras.
Table~\ref{tab:real_tasks} lists the tasks.

\begin{table}[!htbp]
\centering\footnotesize
\caption{\textbf{Real-robot tasks.} A grasp succeeds when the robot holds the target
stably and lifts it clear of its support. Box placement requires the block to stay in the
box, turntable grasp a grasp across the short sides, and ball intercept a stop or deflection
by the gripper. Each task has a 30\,s limit, and timeouts, operator interventions, and
safety stops count as failures.}
\label{tab:real_tasks}
\setlength{\tabcolsep}{4pt}
\begin{tabularx}{\linewidth}{@{}l>{\raggedright\arraybackslash}X@{}}
\toprule
Task & Target motion \\
\midrule
Conveyor grasp & Block on a conveyor belt \\
Moving box placement & Open box on a conveyor belt \\
Turntable grasp & Rectangular block on a turntable \\
Toy car grasp & Toy car driving on the table \\
Ball intercept & Tennis ball rolling at random speed and heading \\
Ramp cart grasp & Block on a cart rolling down a ramp \\
\bottomrule
\end{tabularx}
\end{table}

\paragraph{Data and training.}
\label{app:tracker}
\label{app:lifting}
\label{app:posttraining}
We collect 100 demonstrations per task by teleoperation. FoundationPose
\cite{foundationpose} tracks the target offline in the fixed view, with masks from Grounding DINO \cite{groundingdino} and SAM~2.1 \cite{sam2} and depth
from FoundationStereo \cite{foundationstereo}. Sparse 3D flow comes from CoTracker3
\cite{cotracker3} tracks, lifted to the robot base with stereo depth. The base of \ours
is pretrained on DROID \cite{droid}, with flow labels from the 3D point tracks of PointWorld \cite{pointworld}. DROID has no object poses, so the base has no
target-state tokens and no pose head. For each task, we add them and fine-tune the base on its demonstrations. ACT and $\pi_{0.5}$ are trained on the same
demonstrations with their official recipes.

\paragraph{Deployment.}
\label{app:real_trials}
\label{app:real_results}
At test time, FoundationPose tracks the target online in the fixed view.
Grounding DINO and SAM~2.1 give the first-frame mask, Cutie \cite{cutie} tracks it in
later frames, and FP++ \cite{fpplusplus} gives the initial pose for refinement in each
frame. The tracker and \ours run one after the other on one \cfgGPU{}. The tracker estimates the target pose in each aligned RGB-D frame, and \ours takes this pose and the robot state of the same frame. From the ready frame to the full chunk of eight
commands on the CPU, the latency is 23.92\,ms at the median and 25.17\,ms at p95.
With asynchronous inference, one tracking process and three \ours processes can share
the GPU and reach a throughput of 100.39 frames/s. Fig.~\ref{fig:real_rollouts} shows
a successful trial of \ours and a failed trial of $\pi_{0.5}$ on each task.

\begin{figure*}[p]
\centering
\includegraphics[width=\textwidth]{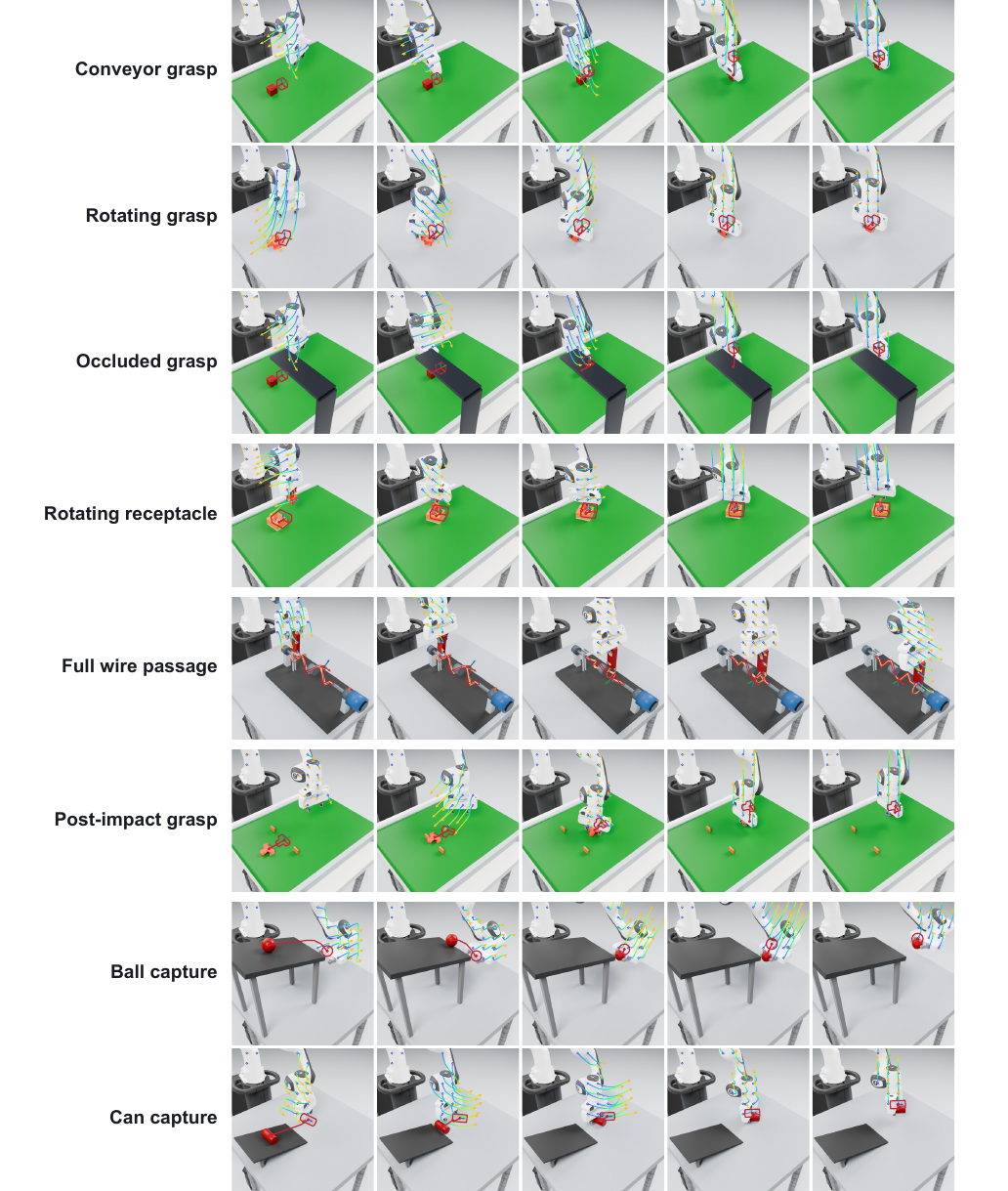}
\caption{\textbf{Rollouts and forecasts of \ours on \bench.} Each row shows five frames
of one rollout. Red shows the forecast path of the target and its forecast pose at the
end, and arrows show the forecast sparse 3D flow.}
\label{fig:rollouts}
\end{figure*}

\begin{figure*}[p]
\centering
\includegraphics[width=\textwidth]{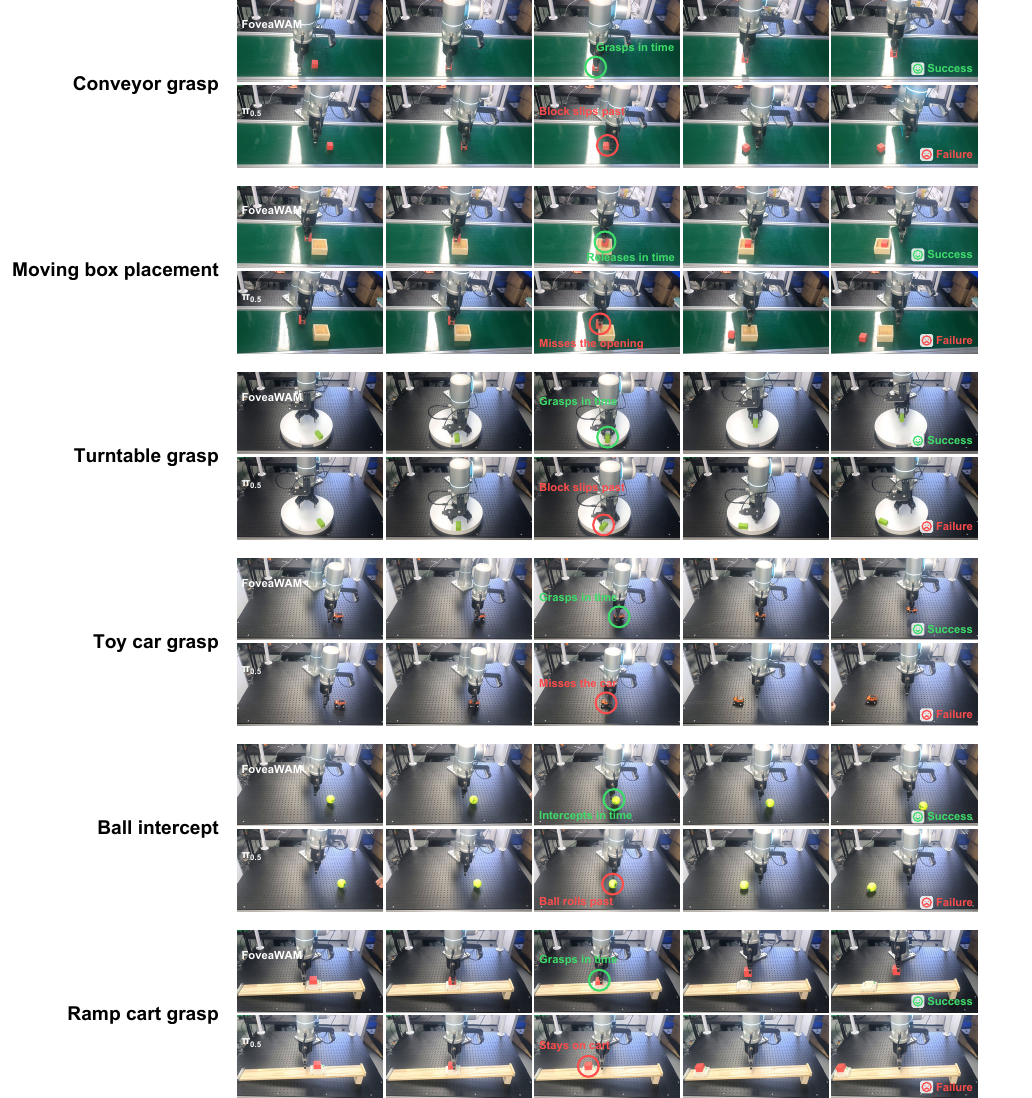}
\caption{\textbf{Real-robot trials.} For each task, the upper row shows a successful
trial of \ours and the lower row a failed trial of $\pi_{0.5}$. Circles mark the targets
in selected frames.}
\label{fig:real_rollouts}
\end{figure*}

\FloatBarrier
\fi
\end{document}